\pdfoutput=1

\documentclass[11pt]{article}

\usepackage[]{acl}

\usepackage{times}
\usepackage{latexsym}

\usepackage[T1]{fontenc}

\usepackage[utf8]{inputenc}

\usepackage{microtype}

\usepackage{tabularx} 
\usepackage{array}
\usepackage{multirow}
\usepackage{booktabs}
\usepackage{graphicx} 
\usepackage{url} 
\usepackage{amsthm,amsmath,amssymb} 
\usepackage{mathrsfs} 
\usepackage{bm} 
\usepackage{pifont} 
\usepackage{float}
\usepackage{lipsum}
\usepackage{soul}
\usepackage{caption}
\usepackage{subcaption}
\usepackage{enumitem}
\usepackage{bbm}
\usepackage{dsfont}
\usepackage{color}
\usepackage{colortbl}
\usepackage{xcolor}
\usepackage{xspace}
\newcommand{\ttbf}[1]{\texttt{#1}}
\newcommand{\name}[0]{\ttbf{FollowBench}\xspace}
\newcommand{\namezh}[0]{\ttbf{FollowBench-zh}\xspace}

\title{\name: A Multi-level Fine-grained Constraints Following Benchmark for Large Language Models}

\author{
Yuxin Jiang$^{1,2}$\thanks{~~Work done during the internship at Huawei Noah's Ark Lab.}, 
Yufei Wang$^{3}$, 
Xingshan Zeng$^{3}$, 
Wanjun Zhong$^{3}$, 
Liangyou Li$^{3}$, 
\\
\textbf{Fei Mi}$^{3}$\textbf{,} 
\textbf{Lifeng Shang}$^{3}$\textbf{,} 
\textbf{Xin Jiang}$^{3}$\textbf{,} 
\textbf{Qun Liu}$^{3}$\textbf{,} 
\textbf{Wei Wang}$^{1,2}$
\\
The Hong Kong University of Science and Technology (Guangzhou)$^1$, \\
The Hong Kong University of Science and Technology$^2$, Huawei Noah’s Ark Lab$^3$ \\
yjiangcm@connect.ust.hk, wang.yufei1@huawei.com, weiwcs@ust.hk 
}

\begin{document}
\maketitle

\begin{abstract}
The ability to follow instructions is crucial for Large Language Models (LLMs) to handle various real-world applications. 
Existing benchmarks primarily focus on evaluating pure response quality, rather than assessing whether the response follows constraints stated in the instruction.
To fill this research gap, in this paper, we propose \name, a \textbf{Multi-level Fine-grained Constraints Following Benchmark} for LLMs. \name comprehensively includes five different types (i.e., Content, Situation, Style, Format, and Example) of fine-grained constraints. To enable a precise constraint following estimation on diverse difficulties, we introduce a \emph{Multi-level} mechanism that incrementally adds a single constraint to the initial instruction at each increased level. To assess whether LLMs' outputs have satisfied every individual constraint, we propose to prompt strong LLMs with constraint-evolution paths to handle challenging open-ended instructions. By evaluating 13 closed-source and open-source popular LLMs on \name, we highlight the weaknesses of LLMs in instruction following and point towards potential avenues for future work. 
The data and code are publicly available at \url{https://github.com/YJiangcm/FollowBench}.
\end{abstract}

\section{Introduction}

\begin{figure}[!tbp]
\centering
\includegraphics[width=\linewidth]{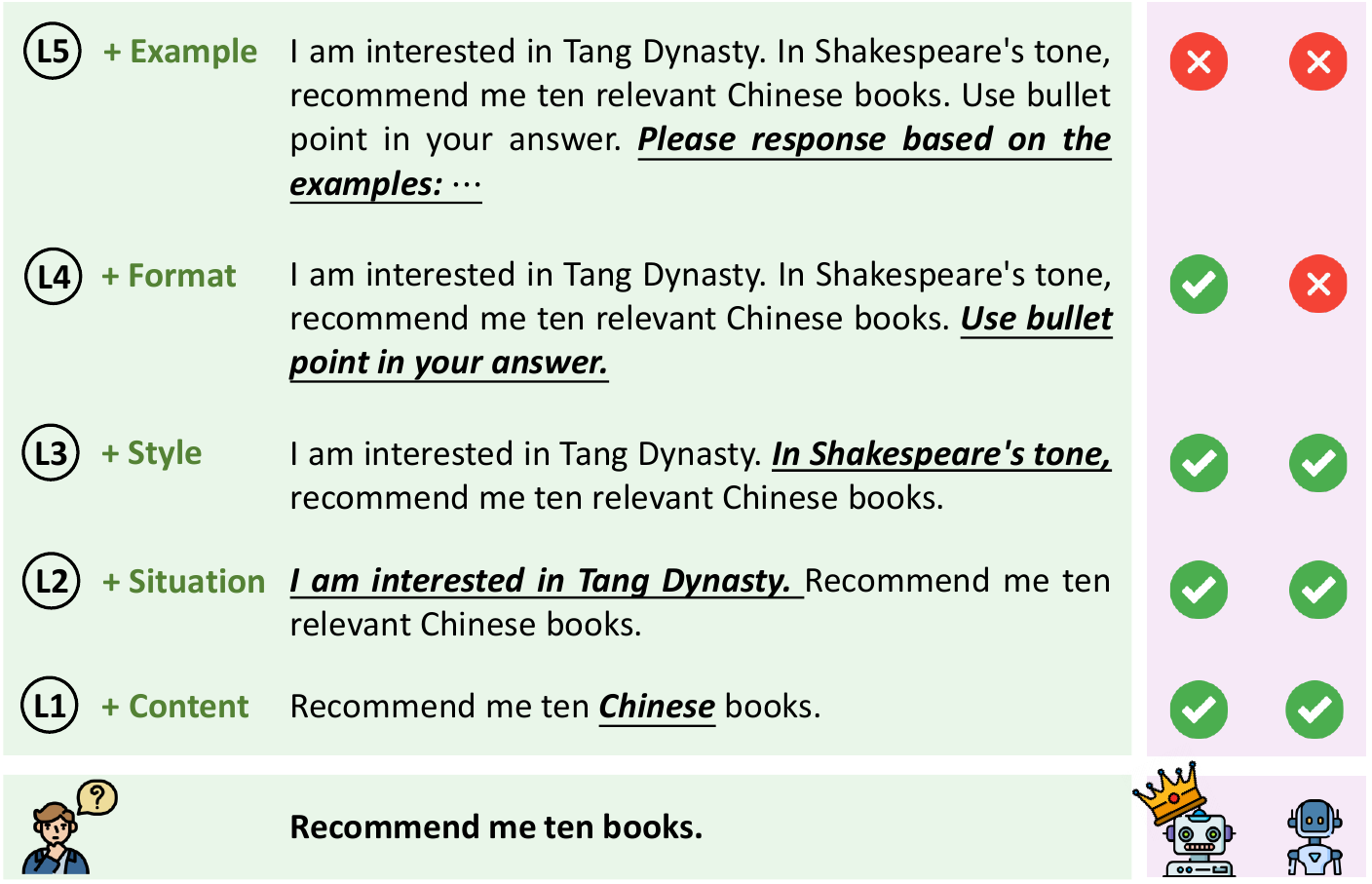}
\caption{
\name covers five \emph{fine-grained} constraint categories and is constructed based on the \emph{Multi-level} mechanism, which increasingly adds a single constraint to straightforward instructions. On the right, the model that can follow instructions with more constraints is deemed to possess better instruction-following ability.
}
\label{fig:followbench}
\end{figure}

Large Language Models (LLMs)~\citep{brown2020language, openai2022chatgpt} pre-trained on web-scale corpora have showcased proficiency in generating fluent and realistic text. Yet, human instructions in real-life cases require the model to generate text that not only possesses a high degree of naturalness but adheres to specific constraints~\cite{yang2023foundation}. For instance, the model may be required to recommend ten books that are specifically written in Chinese (Figure \ref{fig:followbench}), or it might be expected to generate responses that have a certain tone.

The dominant paradigm for assessing if a model can follow instructions involves using human annotators or strongly aligned LLMs to judge its response quality, in terms of helpfulness, relevance, accuracy, depth, creativity, and level of detail~\cite{wang-etal-2023-self-instruct, alpaca_eval, vicuna2023, xu2023wizardlm}. 
However, prior work still has two limitations. Firstly, they ignore the \textbf{fine-grained constraints} inside instructions, which are essential and objective standards for evaluating the instruction-following capability. 
While several benchmarks have rigorously explored individual constraint types, including semantic restrictions~\cite{chen2022controllable} and complex formatting~\cite{tang2023struc}, there exists a lack of comprehensive analysis across the diverse spectrum of constraint categories.
Secondly, few benchmarks consider the varying difficulty of instructions, which is controlled by the number of imposed constraints. This makes it challenging to precisely assess the degree to which LLMs can follow instructions.
Towards this end, our research question is: \emph{how can we systemically and precisely evaluate the instruction-following capability of LLMs?}


In this paper, we construct \name, a \textbf{Multi-level Fine-grained Constraints Following Benchmark}. 
\name comprehensively includes five different types of constraints from real-world scenarios, namely Content (i.e., explicit restrictions on the response content), Situation (i.e., specific situation/background information added to the question), Style (i.e., response style requirements), Format (i.e., response format requirements), and Example (i.e., example pattern recognition and following). 
To precisely estimate the difficulty degree to which LLMs can follow instructions,
as shown in Figure~\ref{fig:followbench}, we propose a novel \emph{Multi-level} mechanism that incrementally adds a single constraint to straightforward instructions at each increased level.
The multi-level mechanism enables us to pinpoint the difficulty level at which LLMs fail to follow instructions, thereby estimating the upper limit of instruction-following capability in LLMs more precisely.
Overall, \name consists of 820 meticulously curated instructions from over 50 NLP tasks, including both closed- and open-ended questions.
For evaluation purposes, we propose a hybrid evaluation method comprising rule-based and model-based solutions. Given LLMs' outputs, both solutions judge whether the outputs satisfy each of the constraints in the instructions.
The rule-based solutions focus on closed-ended instructions while the model-based solutions are applied to opened-ended instructions. For model-based solutions, instead of merely using current instructions and responses as input, we additionally provide the evolution process of the instructions in the input prompts to LLM judges to better understand each individual constraint.  
Both the data construction and the evaluation undergo human verification.

In our experiments, we propose three metrics to assess the instruction-following ability of 13 prominent closed-source and open-source LLMs on \name.
Our principal observations are: (1) the performance of all tested models declines substantially with an increase in difficulty level (the number of constraints in an instruction); (2) although closed-source models such as GPT-4 and GPT-3.5 \textbf{only} consecutively satisfy around three constraints on average, they still markedly surpass all open-source models;
(3) certain specific constraint categories, such as Situation and Example, prove to be more challenging for LLMs than others; (4) beyond capabilities such as knowledge and reasoning, instruction following can offer an additional lens for comprehensively assessing the proficiency of LLMs.

\section{Related Work}

\subsection{Instruction-Following Language Models}

Prior research has found that LLMs fine-tuned with annotated ``instructional'' data, which is composed of language instructional commands and their desired outcomes, can be effective at following general language instructions~\cite{weller-etal-2020-learning, sanh2021multitask, mishra-etal-2022-cross, DBLP:journals/corr/abs-2402-11905}. 
To enhance the understanding of LLMs regarding the intricate and varied intentions of users in real-world scenarios, works like ChatGPT~\cite{openai2022chatgpt} and GPT-4~\cite{DBLP:journals/corr/abs-2303-08774} implement instruction tuning across a wide range of human-crafted instructions and task categories.
Recent studies~\cite{vicuna2023, xu2023wizardlm, DBLP:conf/emnlp/JiangCCW23} have pivoted towards automatically generating high-quality data to enhance the instruction-following capability of LLMs, addressing the challenges posed by labor-intensive human annotation.

\subsection{Evaluation for Instruction Following}
There are several research efforts in evaluating LLMs' following capability towards particular tasks. 
~\citet{tang2023struc} focuses on evaluating LLMs' generation capability towards complex structured tabular data in text, HTML, and Latex. They first collect tables from existing NLP benchmarks and websites, then construct guiding instructions based on these data.
~\citet{chen2022controllable} evaluates whether LLMs can follow particular knowledge-intensive generation instructions. They first provide a list of examples (e.g., a list of sports stars in the UK), followed by a constraint that is contradicted by the examples (e.g., not mentioning any athletes). These benchmarks can only demonstrate particular types of instruction-following capability of LLMs.
In contrast, \name comprehensively includes instructions with five different types of fine-grained constraints in multi-level difficulty and \name should provide a well-rounded and precise estimation of instruction-following capability for existing LLMs. For more details on LLM evaluation, we refer to recent surveys~\cite{chang2023survey,wang2023aligning}.

\begin{table*}[]
\footnotesize
\centering
\begin{tabular}{llccc}
\toprule
\textbf{Constraint}        & \textbf{Task}                                        & \textbf{Avg Len} & \textbf{\#Data} & \textbf{Evaluation} \\ \midrule
                           & Data-to-Text Generation                              & 84               & 25              & \includegraphics[width=0.25cm]{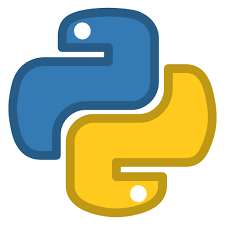}                    \\
                           & Document-Level Event Argument Extraction             & 696              & 25              & \includegraphics[width=0.25cm]{figures/python.png}                    \\
                           & Document-Level Named Entity Recognition            & 376              & 25              &  \includegraphics[width=0.25cm]{figures/python.png}                   \\
                           & Text Generation with Language Constraints            & 88               & 25              & \includegraphics[width=0.25cm]{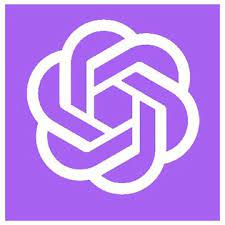}                    \\
\multirow{-5}{*}{Content}  & Open-ended Question Answering                        & 56               & 25              &  \includegraphics[width=0.25cm]{figures/gpt2.jpg}   \\ \midrule
                           & Suggestion Generation & 69               & 40              & \includegraphics[width=0.25cm]{figures/gpt2.jpg}   \\
                           & Role-playing & 111               & 15              & \includegraphics[width=0.25cm]{figures/gpt2.jpg}                      \\
\multirow{-3}{*}{Situation} & Complex Situation Reasoning                                   & 102               & 55              & \includegraphics[width=0.25cm]{figures/python.png} \\  \midrule
Style                      & Open-ended Question Answering & 64               & 150             & \includegraphics[width=0.25cm]{figures/gpt2.jpg}  \\ \midrule
                     & Text-to-Table Generation & 171               & 30             & \includegraphics[width=0.25cm]{figures/python.png} \\
                           \multirow{-3}{*}{Format} & Open-ended Question Answering & 74               & 120             & \includegraphics[width=0.25cm]{figures/gpt2.jpg}  \\ \midrule
Example                    & 40 diverse NLP tasks                                 & 739              & 200             & \includegraphics[width=0.25cm]{figures/python.png} \\ \midrule
                           & Text Editing & 96               & 25              & \includegraphics[width=0.25cm]{figures/python.png} \\
                           & Summarization                                        & 254              & 25              & \includegraphics[width=0.25cm]{figures/python.png}                    \\
                           & Machine Translation                                  & 91               & 25              & \includegraphics[width=0.25cm]{figures/python.png}                    \\
\multirow{-4}{*}{Mixed}    & Story Generation                                     & 34               & 10              & \includegraphics[width=0.25cm]{figures/gpt2.jpg}    \\ \bottomrule
\end{tabular}
\caption{An overview of \name. ``Avg Len'' is the average word number of instructions. \includegraphics[width=0.25cm]{figures/python.png} refers to rule-based evaluation, while \includegraphics[width=0.25cm]{figures/gpt2.jpg} refers to model-based evaluation.}
\label{tab:stat}
\end{table*}

\section{\name}
\label{sec: followbench}
As shown in Table \ref{tab:stat}, \name encompasses five distinct \emph{fine-grained} constraint categories: Content, Situation, Style, Format, and Example. Each category consists of instructions from various NLP tasks. 
Different from previous benchmarks, we introduce a \emph{Multi-level} mechanism that incrementally adds constraints to an initial instruction (see examples in Figure \ref{fig:intro}), producing a set of instructions ranging from 1 to 5 constraints. 
In the following part of this paper, we use ``level $n$'' to denote an instruction containing $n$ constraints.
It is worth noticing that the way of adding constraints is meticulously designed for each task within its respective constraint category.
The multi-level mechanism enables us to pinpoint the difficulty level at which LLMs fail to follow instructions, thereby estimating the upper bound of instruction-following capability in LLMs more precisely.

To encapsulate, we will introduce the data construction process of \name, including \emph{fine-grained} constraints and the \emph{Multi-level} mechanism, in \S\ref{sec: data_construction}. 
In \S\ref{sec: evaluation}, we propose an evaluation protocol with three metrics that seamlessly integrate with the multi-level mechanism.




\subsection{Data Construction}
\label{sec: data_construction}

\begin{figure*}[!t]
\centering
\includegraphics[width=0.9\linewidth]{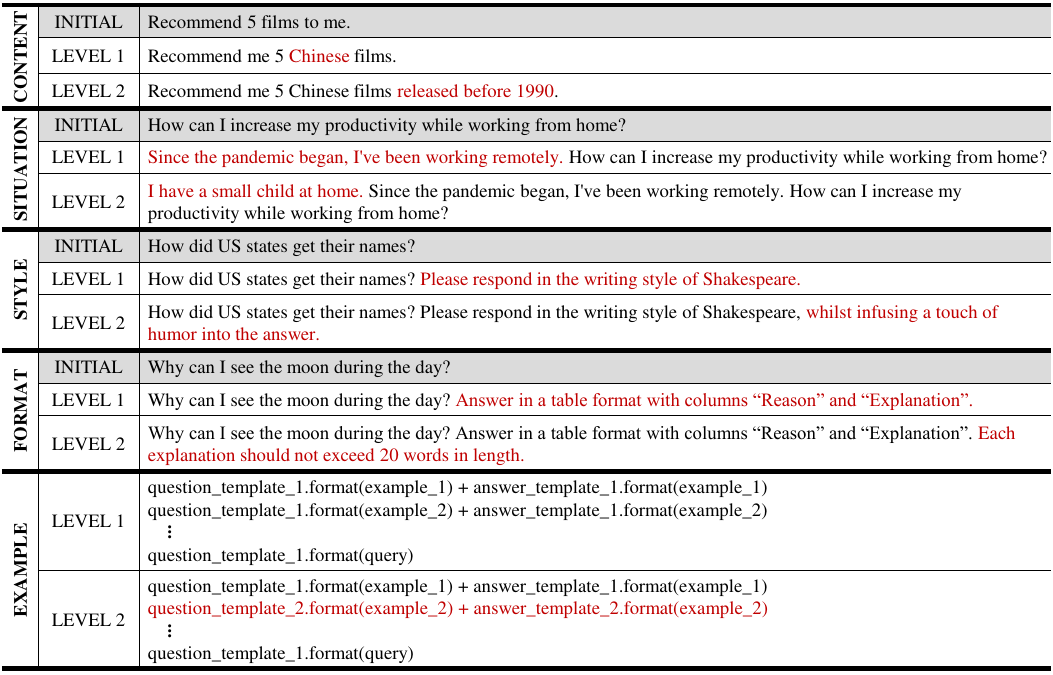}
\caption{
\name covers five \emph{fine-grained} categories of constraints. Within each constraint type, we construct a range of \emph{Multi-level} instructions by incrementally adding constraints (highlighted in red). There are five levels in total; however, we only display the first two levels from each category for demonstration purposes.
}
\label{fig:intro}
\end{figure*}

\paragraph{Content Constraints}
Content constraints refer to \emph{explicit} impositions of specific conditions that shape the depth or scope of the response content.
An example is shown in Figure \ref{fig:intro}, which sets specific criteria for the retrieved object.
Ensuring that LLMs adhere to content constraints has become a critical challenge in Controlled Text Generation~\cite{zhang2022survey}, as it demands models to understand specific guidelines and adapt responses to prescribed conditions~\cite{chen2022controllable}.
To this end, we first collect data from the following tasks: (1) Complex Information Extraction aims at retrieving specific information about specific objects from the given text; (2) Text Generation with Language Constraints requires to generate fluent on-topic content while respecting a specified constraint; (3) Open-ended Question Answering comes from real scenarios (e.g., open-source platforms) to prevent the risk of data leakage. 
Subsequently, we construct multi-level instructions by adding one content constraint to the collected instructions each time. The manners of introducing additional constraints depend on different tasks (see details in Appendix \ref{appendix: data_generation_content}).
For Complex Information Extraction, we gradually narrow down the scope of the information to be extracted.
For Text Generation with Language Constraints, we incorporate additional restrictions from WordNet~\cite{miller-1992-wordnet} and Wikidata~\cite{vrandevcic2014wikidata}. 
For Open-ended Question Answering, we utilize advanced LLMs like GPT-4 to generate a new instruction with one more constraint based on the given instruction.
While the output from the LLMs serves primarily as a reference, we handpick the most relevant and challenging synthesized instructions to ensure data quality.

\paragraph{Situation Constraints}
Situation Constraints refer to impositions of specific situations or backgrounds that \emph{implicitly} guide the appropriate answer of the response.
For instance, it is necessary to illustrate the situation when asking for customized suggestions, as shown in Figure \ref{fig:intro}.
Another example is to customize LLMs to simulate various characters under certain circumstances, namely Role-playing, which provides a more nuanced interaction for users~\cite{shanahan2023role, wang2023interactive}.
Situation constraints push LLMs beyond mere factual retrieval or surface-level synthesis, demanding a nuanced understanding, a dynamic adaptation, and complicated reasoning to the situation~\cite{yao2022react, liu2023agentbench}.
Besides real-life questions, we also consider Complex Situation Reasoning tasks including Math Word Problems, Time/Spatial Reasoning, and Code Generation. These tasks all require interpreting and solving problems within a given situation,
thus matching the definition of situation constraints.
We first collect initial instructions from these sources and then manually curate multi-level instructions by incrementally supplementing situation information inside (see Appendix \ref{appendix: data_generation_scenario}).



    


\paragraph{Style Constraints}
Style Constraints control the stylistic variations of output to accomplish specific stylistic goals, such as tone, sentiment, formality, and empathy~\cite{tsai-etal-2021-style}, as illustrated in Figure \ref{fig:intro}.
The challenges of style constraints for LLMs are the intricate understanding and adaptation of language nuances, ensuring contextually appropriate and stylistically consistent outputs~\cite{smith2020controlling, cheng2022replacing}.
Drawing from Open-ended Question Answering datasets and online platforms, we collect initial instructions and then leverage LLMs' in-context learning capability to craft instructions with multi-level style constraints. The prompt template can be viewed in Figure \ref{fig:style_prompt}. Human experts subsequently review and refine the outputs produced by LLMs.

\paragraph{Format Constraints}

Format Constraints refer to stipulations governing the structural, linguistic, or output presentation of generated content.
An example is shown in Figure \ref{fig:intro}, which sets limits on word length and requires the format of the response to be a table.
Format constraints necessitate a deep, nuanced understanding of language and structure, allowing them to flexibly adapt outputs according to diverse and often intricate specifications~\cite{zhao2023large}.
Recent work has pointed out that even the most superior LLMs may struggle with tasks that require generating complex, structured outputs such as tables, JSON, HTML, or LaTeX~\cite{tang2023struc}.
To include a variety of format constraints, we first collect instructions from broader domains, encompassing Text-to-Table Generation and Open-ended Question Answering, then we utilize powerful LLMs to sequentially add format constraints ranging from length and hierarchy to specialized linguistic features and output mediums. See Figure \ref{fig:format_prompt} for the prompt template.
Finally, we ask human experts to carefully check and refine the synthesized instructions.

\paragraph{Example Constraints}
LLMs have demonstrated stunning few-shot learning ability~\cite{brown2020language}, which enables them to adapt quickly to a new query by recognizing patterns from just a few examples provided in the prompt.
However, the robustness of few-shot learning, which means whether LLMs can still follow correct patterns after introducing ``noise'' examples, has not been explored.
Thus, we propose a novel constraint category named Example Constraints to evaluate the example pattern recognition and following capability of LLMs.
We automatically craft instructions with multi-level example constraints based on PromptSource~\cite{bach-etal-2022-promptsource}, where instructions at level $n$ have $n-1$ noise examples in the input. The details are illustrated in Appendix \ref{appendix: data_generation_example}.


\paragraph{Mixed Constraints}
For the above five constraint categories, we construct multi-level instructions by adding the same type of constraint sequentially.
Nevertheless, real-world scenarios often require more than one type of constraint to be enforced in a singular instruction.
Therefore, we define Mixed Constraints as the composition of varied constraint categories.
For instance, in the Text Editing task, we may want to add some content as well as adjust the output format.
Besides, we also consider several tasks that are naturally suitable for constructing mixed constraints, including Summarization, Machine Translation, and Story Generation (see Appendix \ref{appendix: data_generation_mixed}).
Instructions with multi-level mixed constraints are produced by specifying the format of generating answers (Format Constraints), requiring the generated text to include or not include certain keywords (Content Constraints), etc.

\paragraph{Data Quality Control}
To ensure the data quality of \name, we implement a dual-layer verification system for each instruction. Two annotators independently evaluate: (1) the appropriateness of the instruction for its designated constraint category, and (2) the validity of the added constraint within the instruction. In instances of divergent evaluations, a third annotator intervenes for a detailed review to ensure consensus.

\begin{figure}[!t]
\centering
\includegraphics[width=\linewidth]{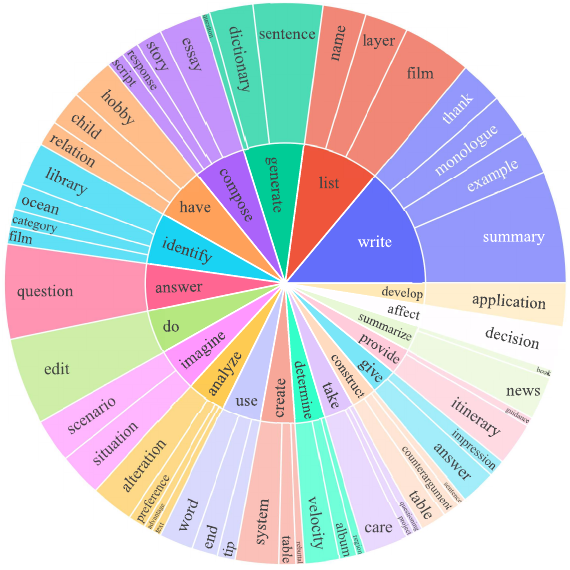}
\caption{
Verb-noun structure of \name Instructions.
}
\label{fig:verb_noun}
\end{figure}

We analyze the comprehensiveness and diversity of in \name, which includes 820 instructions in total.
To maintain data diversity, we strive to ensure that the ROUGE-L score between any two initial instructions is below 0.7.
Figure \ref{fig:verb_noun} shows the verb-noun structure of \name instructions, where the top 20 verbs (inner circle) and their top 4 direct noun objects (outer circle) are depicted.

\subsection{Evaluation Protocol}
\label{sec: evaluation}

\begin{figure}[!tbp]
\centering
\includegraphics[width=\linewidth]{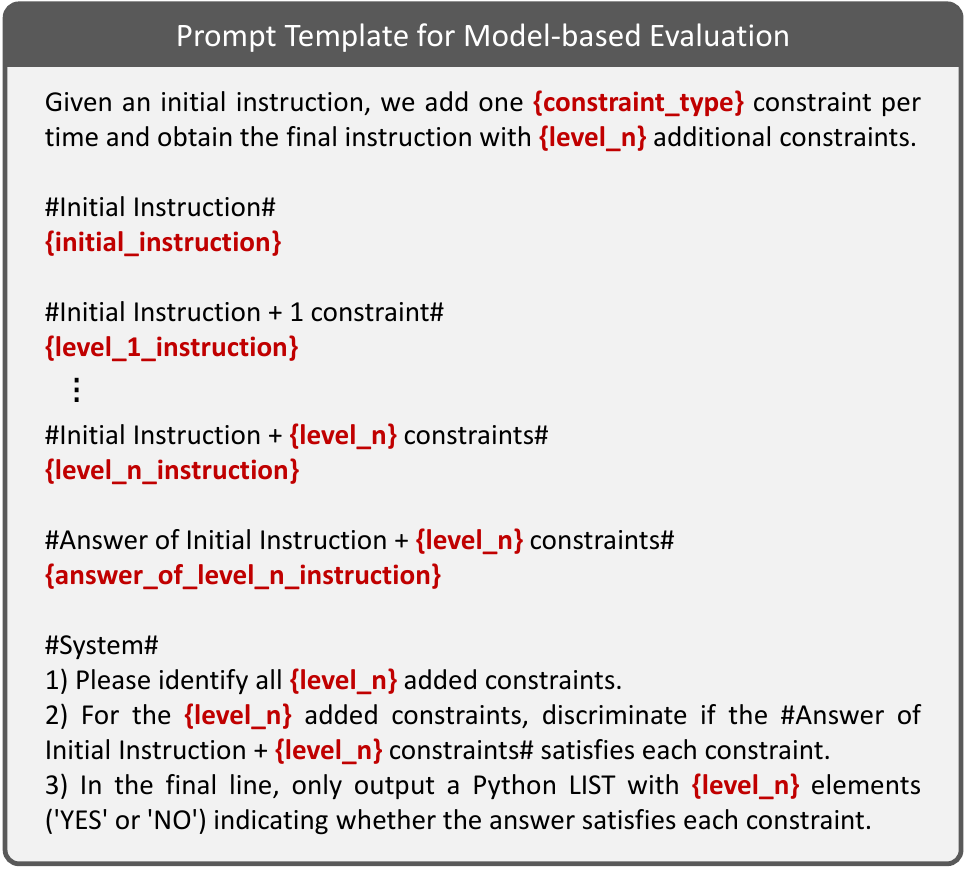}
\caption{
Prompt template for model-based evaluation.
}
\label{fig:eval_prompt}
\end{figure}

Given that nearly half of instructions in \name are open-ended without reference answers, devising a rule-based program to assess the outputs is extremely challenging.
To overcome this, inspired by~\cite{gilardi2023chatgpt, huang2023chatgpt}, we propose to develop a model-based approach by using strong LLMs\footnote{We use GPT-4-Preview-1106 in our experiments.} as judges.
Previous works leverage strong LLMs to determine the quality of a response, by prompting them to consider multiple factors such as usefulness, relevance, and level of detail~\cite{alpaca_eval, vicuna2023}.
To effectively guide strong LLMs to judge the constraint following capability objectively and faithfully, we propose a \emph{Multi-level-aware} prompt template, as shown in Figure \ref{fig:eval_prompt}.
Rather than merely presenting the instruction and asking LLMs to determine whether all constraints are satisfied, we illustrate the evolution process of the instruction and prompt LLMs to pinpoint the newly added constraint at each level.
Exposing the evolution process of the instruction allows for a more granular understanding and identification of individual constraints, enhancing LLMs' ability to discriminate with precision. The ablation study in \S\ref{sec: high_agreement_with_human} validates the effectiveness of this strategy.

Moreover, we propose three novel metrics to evaluate the instruction-following ability of LLMs. 
For an instruction with $n$ constraints (level $n$), we use the rule-based program or LLM judge (refer to Table \ref{tab:stat}) to discriminate if the response of a model satisfies each constraint in the instruction. 
At each level $n$, given a set of $m$ instructions, we define the Hard Satisfaction Rate (HSR) and Soft Satisfaction Rate (SSR) as follows:
\begin{align}
\text{HSR} &= \frac{1}{m} \sum_{i=1}^m \prod_{j=1}^n s_i^j \\
\text{SSR} &= \frac{1}{mn} \sum_{i=1}^m \sum_{j=1}^n s_i^j
\label{equation: ssr}
\end{align}
where $s_i^j=1$ if the $j$-th constraint of $i$-th instruction is satisfied and $s_i^j=0$ otherwise.
HSR measures the average rate at which all constraints of individual instructions are fully satisfied, while SSR calculates the average satisfaction rate of individual constraints across all instructions. 

As described in \S\ref{sec: followbench}, we construct \name by incrementally adding five constraints to an initial instruction, enabling us to pinpoint the difficulty level at which LLMs fail to follow instructions.
Therefore, we propose a metric called Consistent Satisfaction Levels (CSL) to estimate how many consecutive levels a model can satisfy, beginning from level 1:
\begin{align}
\text{CSL} = \frac{1}{g} \sum_{i=1}^g \mathop{\arg\max}_{l} \left(l \times \prod_{n=1}^l S_{i}^n\right)
\label{equation: csl}
\end{align}
where $g$ is the group number of initial instructions, $S_{i}^n=1$ if all constraints of the $i$-th instruction at level-$n$ are satisfied and $S_{i}^n=0$ otherwise.

\section{Experiments}
This section first introduces experimental setup in \S\ref{sec: experimental_setup}, and then presents the main experiment results across two key dimensions: difficulty level in \S\ref{sec: level_categorized_results} and constraint category in \S\ref{sec: constraint_categorized_results}.

\subsection{Experimental Setup}
\label{sec: experimental_setup}
We evaluate 13 popular LLMs 
including GPT-4-Preview-1106~\cite{DBLP:journals/corr/abs-2303-08774}, GPT-3.5-Turbo-1106~\cite{openai2022chatgpt}, Qwen-Chat-72B/14B/7B~\cite{qwen}, LLaMA2-Chat-70B/13B/7B~\cite{touvron2023llama}, WizardLM-13B-V1.2~\cite{xu2023wizardlm}, Vicuna-13B/7B-V1.5~\cite{vicuna2023}, Baichuan2-Chat-7B~\cite{baichuan2023baichuan2}, and ChatGLM3-6B~\cite{du2022glm}.
We access GPT-4-Preview-1106 and GPT-3.5-Turbo-1106 via OpenAI API. 
We access other open-source LLMs from their official repositories.
During the inference process, we set the temperature to 0 to ensure deterministic outputs. We set the maximum generation length to 2048. Other parameters use their default values.
To facilitate the multilingual evaluation of LLM’s instruction-following ability, we also craft a Chinese version of \name, namely \namezh, in Appendix \ref{appendix: followbench_zh}.

\subsection{Level-categorized Results}
\label{sec: level_categorized_results}

\definecolor{colorGPT}{rgb}{0.95, 0.95, 1.0}    
\definecolor{color70B}{rgb}{0.95, 1.0, 0.95} 
\definecolor{color13B}{rgb}{1.0, 0.95, 0.95}  
\definecolor{colorOther}{rgb}{0.95, 0.95, 0.95} 

\begin{table*}[]
\small
\centering
\setlength{\tabcolsep}{6pt} 
\begin{tabular}{l|cccccc|cccccc|c}
\toprule
& \multicolumn{6}{c|}{\textbf{HSR (\%)}} & \multicolumn{6}{c|}{\textbf{SSR (\%)}} \\
\cmidrule(lr){2-7} \cmidrule(lr){8-13} 
\multirow{-2}{*}{\centering\textbf{Model}}
& \textbf{L1} & \textbf{L2} & \textbf{L3} & \textbf{L4} & \textbf{L5} & \textbf{Avg.} & \textbf{L1} & \textbf{L2} & \textbf{L3} & \textbf{L4} & \textbf{L5} & \textbf{Avg.}
& \multirow{-2}{*}{\centering\textbf{CSL}} \\ 
\midrule
\rowcolor{colorGPT}
GPT-4-Preview-1106 & \textbf{84.7} & \textbf{75.6} & \textbf{70.8} & \textbf{73.9} & \textbf{61.9} & \textbf{73.4} & \textbf{84.7} & \textbf{77.0} & \textbf{75.3} & \textbf{77.0} & \textbf{72.3} & \textbf{77.2} &\textbf{3.3} \\
\rowcolor{colorGPT}
GPT-3.5-Turbo-1106 &80.3 &68.0 &68.6 &61.1 &53.2 &66.2 &80.3 &71.2 &74.2 &69.6 &67.1 &72.5 &2.9 \\
\rowcolor{color70B}
Qwen-Chat-72B & 73.8 & 63.3 & 54.3 & 45.2 & 39.9 &55.3 & 73.8 & 67.5 & 63.2 & 57.6 & 56.0 &63.6 &2.4 \\
\rowcolor{color70B}
LLaMA2-Chat-70B & 59.9 & 53.3 & 46.0 & 40.2 & 37.9 &47.5 & 59.9 & 57.3 & 55.7 & 53.3 & 53.2 &55.9 &2.1 \\
\rowcolor{color13B}
Qwen-Chat-14B & 62.8 & 56.2 & 47.7 & 38.7 & 30.9 &47.3 & 62.8 & 61.9 & 57.7 & 52.6 & 51.4 &57.3 &1.9 \\
\rowcolor{color13B}
WizardLM-13B-V1.2 & 68.8 & 64.1 & 53.1 & 40.8 & 35.8 &52.5 & 68.8 & 65.7 & 61.8 & 53.4 & 53.9 &60.7 &2.2 \\
\rowcolor{color13B}
LLaMA2-Chat-13B & 57.0 & 56.0 & 50.4 & 44.4 & 38.1 &49.2 &57.0&60.0&58.0&54.8&52.2 &56.4 &2.2 \\
\rowcolor{color13B}
Vicuna-13B-V1.5 & 71.2 & 60.2 & 49.6 & 40.6 & 34.0 &51.1 &71.2&64.8&59.9&54.5&53.6 &60.8 &2.1 \\
\rowcolor{colorOther}
Qwen-Chat-7B & 55.9 & 51.7 & 38.7 & 33.1 & 23.3 &40.6 & 55.9 & 58.2 & 51.6 & 48.9 & 45.9 &52.1 &1.5 \\
\rowcolor{colorOther}
LLaMA2-Chat-7B & 58.0 & 51.3 & 47.4 & 39.5 & 35.3 &46.3 & 58.0&56.5&55.6&52.5&51.4 &54.8 &1.9\\
\rowcolor{colorOther}
Vicuna-7B-V1.5 & 60.8 & 52.0 & 42.2 & 33.3 & 23.9 &42.4 &60.8&58.6&55.5&48.3&49.0 &54.4 &1.7 \\
\rowcolor{colorOther}
Baichuan2-Chat-7B &58.3 &46.1 &40.7 &30.4 &25.5 &40.2 &58.3 &55.4 &54.9 &49.9 &49.3 &53.6 &1.4 \\
\rowcolor{colorOther}
ChatGLM3-6B &60.9 &46.6 &36.7 &27.8 &21.4 &38.7 &60.9 &55.3 &51.2 &47.9 &45.0 &52.0 &1.6 \\
\bottomrule
\end{tabular}
\caption{Results across five difficulty levels. For each level, we compute the average score of all constraint categories. \colorbox{colorGPT}{Proprietary LLMs}, \colorbox{color70B}{open-sourced LLMs (large)}, \colorbox{color13B}{open-sourced LLMs (medium)}, and \colorbox{colorOther}{open-sourced LLMs (small)} are distinguished by different colors.}
\label{tab:level}
\end{table*}

Table \ref{tab:level} provides a comprehensive comparison of various models across five difficulty levels, denoted as L1 to L5. The detailed results for each constraint category are listed in Appendix \ref{appendix: experiments}.
From a bird's-eye view, we can infer that the performance typically diminishes as we progress from L1 to L5 for almost all models. This trend coincides with the increasing complexity or stringent requirements associated with higher levels.
Besides, models with larger architectures generally outperform their smaller counterparts. However, it's worth noting that the scaling law does not apply as effectively to LLaMA2-Chat-70B.
It can be observed from Appendix \ref{appendix: experiments} that while LLaMA-2-Chat-70B does indeed outperform LLaMA-2-Chat-13B in Situation constraints, it shows a relative underperformance in Format and Mixed Constraints categories.
More importantly, there's a marked performance gap between closed-source models (i.e., GPT-4 and GPT-3.5) and open-source models.
Regarding CSL, it can be deduced that the instruction-following upper bound for GPT-4 and GPT-3.5 is approximately 3 constraints (level 3) added to an initial instruction. In contrast, open-source models typically have an upper limit of about 2 constraints (level 2).
This significant difference underscores the better instruction-following ability of proprietary models, possibly due to superior data quality or optimization strategies such as RLHF~\cite{ouyang2022training}.
Furthermore, \textbf{even the most sophisticated models are limited to following instructions with about three constraints}, suggesting significant potential for further improvement.

\subsection{Constraint-categorized Results}
\label{sec: constraint_categorized_results}

\begin{figure}[!tbp]
\centering
\includegraphics[width=\linewidth]{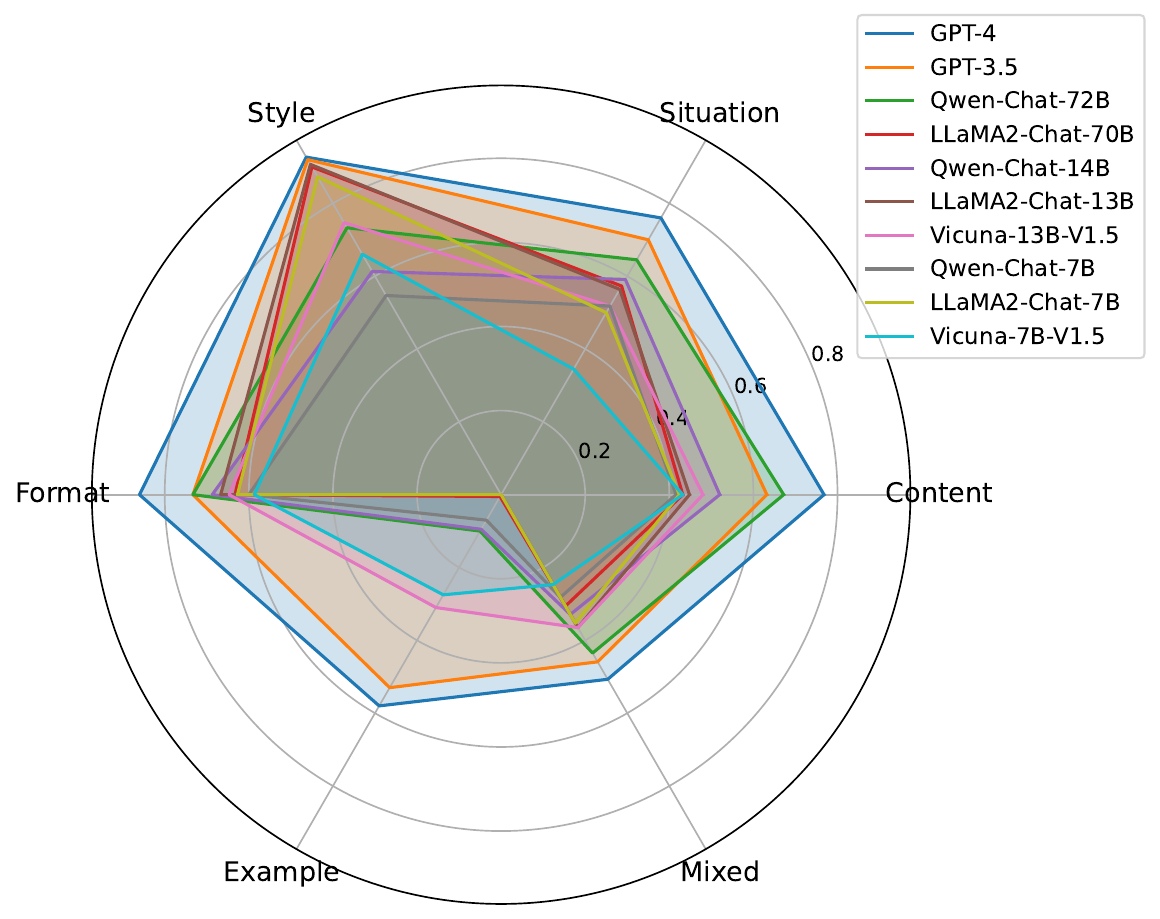}
\caption{
HSR (\%) results in diverse constraint categories. For each category, we compute the average score of all difficulty levels.
}
\label{fig:category}
\end{figure}

As depicted in Figure \ref{fig:category}, we assess various models over different constraint categories to succinctly showcase the instruction-following capability of LLMs in a singular dimension.
Notably, GPT-4 and GPT-3.5 surpass open-source models in every constraint category, with a pronounced advantage in Content, Situation, Example, and Mixed constraints.
Furthermore, most models demonstrated commendable proficiency under the Style constraint. While GPT-4, GPT-3.5, and LLaMA2-Chat-70B were the frontrunners, the trend suggests that style adaptation is an area where many models excel, hinting at its utility in real-world applications.
However, the Example and Mixed constraints posed a challenge to most models. While GPT-4 led the segment, even its scores were noticeably lower than in other categories.
To illustrate, in the ``Example'' category, we evaluated the instruction-following capabilities of LLMs by introducing ``noise examples'' with varying natural language templates. The observed performance decline is primarily due to the LLMs' limited training in processing such noisy inputs within context-based learning scenarios. Typically, LLMs are fine-tuned on clean and uniform datasets, which do not adequately prepare them to sift through and ignore irrelevant or misleading information. This limitation becomes apparent when faced with the intricacies of real-world data.
Our findings underscore the complexity of these constraints and pinpoint an area for potential improvement.

\section{Analysis}
This section includes: an ablation study confirming our prompt template's effectiveness for model-based evaluation (\S\ref{sec: high_agreement_with_human}); a comparison of instruction following vs. other LLM's abilities (\S\ref{sec: instruction_following_vs_others}); an examination of failure consistency (\S\ref{sec: failure_consistency}); and an investigation of various decoding strategies (\S\ref{sec: decoding_strategies}).
In addition, a case study is presented in Appendix \ref{appendix: case_study} for further analysis.

\subsection{Ablation Study of Model-based Evaluation}
\label{sec: high_agreement_with_human}

We randomly sample 100 cases that require LLM evaluation, encompassing five constraints, five distinct levels, and four diverse models to guarantee comprehensive representation.
Then we ask three expert-level human labelers to assess whether the model's response satisfies all the constraints in each case and use the majority voting as the final human annotations.
As shown in Table \ref{tab:human_agreement}, our prompt template (Figure \ref{fig:eval_prompt}) registers an impressive 88\% agreement with expert human evaluations, surpassing even the internal agreement among human experts, which stands at 85\%.
Remarkably, when the evolution process of multi-level constraints is removed from our prompt template, the agreement rate dips by 9\%. This underlines the instrumental role played by the detailed portrayal of the instruction's evolution in enhancing LLM's precision in discernment.
In contrast, we also employ the prompt template from Vicuna~\cite{vicuna2023}, a standard prompt for assessing the overall quality of response. This template prompts the LLM to assign a score from 0 to 10 for each response. 
We consider responses with a score above 5.0 to meet all the constraints of an instruction. This approach achieves 67\% agreement with human evaluators. Such a disparity highlights the fundamental difference between assessing the instruction-following ability and the overall response quality.

\begin{table}[t]
\small
\centering
\begin{tabular}{lc}
\toprule
\textbf{Prompt}    & \textbf{Agreement with Human} \\
\midrule
Ours             &       \textbf{88\%}                 \\
Ours w/o ML &          79\%              \\
Vicuna-Single    &       67\%                  \\  
\bottomrule
\end{tabular}
\caption{Agreement between human and diverse prompt templates. We use ML to denote multi-level.}
\label{tab:human_agreement}
\end{table}

\subsection{Instruction Following vs. Other Abilities}
\label{sec: instruction_following_vs_others}

Table \ref{tab:different_benchmark} presents a comparison of representative LLMs across different abilities, not just instruction following (\name). This includes overall response quality (AlpacaEval~\cite{alpaca_eval}), knowledge (MMLU~\cite{hendryckstest2021}), and reasoning (BBH~\cite{suzgun2022challenging}).
We can find that our \name provides an additional perspective for a holistic LLM evaluation.
As an illustration, while the performance of WizardLM-13B-V1.2 exceeds that of GPT-3.5 in terms of overall response quality, it notably lags behind in instruction-following ability.
Similarly, Vicuna-V1.5 excels over LLaMA2-Chat in the realms of knowledge and reasoning but struggles with instruction-following tasks.

\begin{table}[t]
\tiny
\centering
\begin{tabular}{lcccccc}
\toprule
\textbf{Model} & \textbf{Following} & \textbf{Overall} &\textbf{Knowledge} & \textbf{Reasoning} \\
\midrule
GPT-4-Preview-1106                                  & 3.3          & 97.7          & 86.4  & 86.7        \\
GPT-3.5-turbo-1106                                                       & 2.9          & 86.3              & 70.0  & 70.1        \\
LLaMA2-Chat-70B & 2.1 & 92.7 & 63.0 & 60.8 \\
WizardLM-13B-V1.2                                         & 2.2          & 89.2                            & 52.7   & --      \\
LLaMA2-Chat-13B                                         & 2.2          & 81.1                               & 53.6   & 40.2       \\
Vicuna-13B-V1.5                                            & 2.1          & --                                 & 55.8  & 51.5        \\
LLaMA2-Chat-7B                                     & 1.9          & 71.4                              & 45.8   & 35.6       \\
Vicuna-7B-V1.5                              & 1.7          & --                                & 49.8  & 43.4        \\
\bottomrule
\end{tabular}
\caption{Model comparison on different abilities.}
\label{tab:different_benchmark}
\end{table}

\subsection{Does Failure at Lower Level Necessarily Lead to Failure at Higher Level?}
\label{sec: failure_consistency}

\begin{table}[t]
\small
\centering
\begin{tabular}{lc}
\toprule
\textbf{Model}    & \textbf{Failure Consistency (\%)} \\
\midrule
GPT-4-Preview-1106 & 42.2                        \\
WizardLM-13B-V1.2 & 57.3                        \\
Vicuna-7B-V1.5    & 61.8                        \\
ChatGLM3-6B       & 64.0         \\   
\bottomrule
\end{tabular}
\caption{Results on failure consistency.}
\label{tab:failure_consistency}
\end{table}

For a set of instructions that has five difficulty levels, if a model's response doesn't satisfy the constraints at level $n$, where $n$ ranges from 1 to 4, we define the \textit{failure consistency} as the percentage that the response will also not fulfill the constraints at any subsequent level greater than $n$.
Combining Table \ref{tab:level} and Table \ref{tab:failure_consistency}, it can be seen that models with better instruction-following capability may exhibit lower failure consistency. 
One possible reason is that the instruction-following ability of more powerful models is less sensitive to the number of constraints in an instruction, thus they are better equipped to adapt and fulfill the requirements even as the constraints increase. This adaptability means that while they may falter at a lower difficulty level, they can still manage to meet the demands of higher difficulty levels, leading to a decrease in failure consistency.

\begin{figure}[!tbp]
\centering
\includegraphics[width=\linewidth]{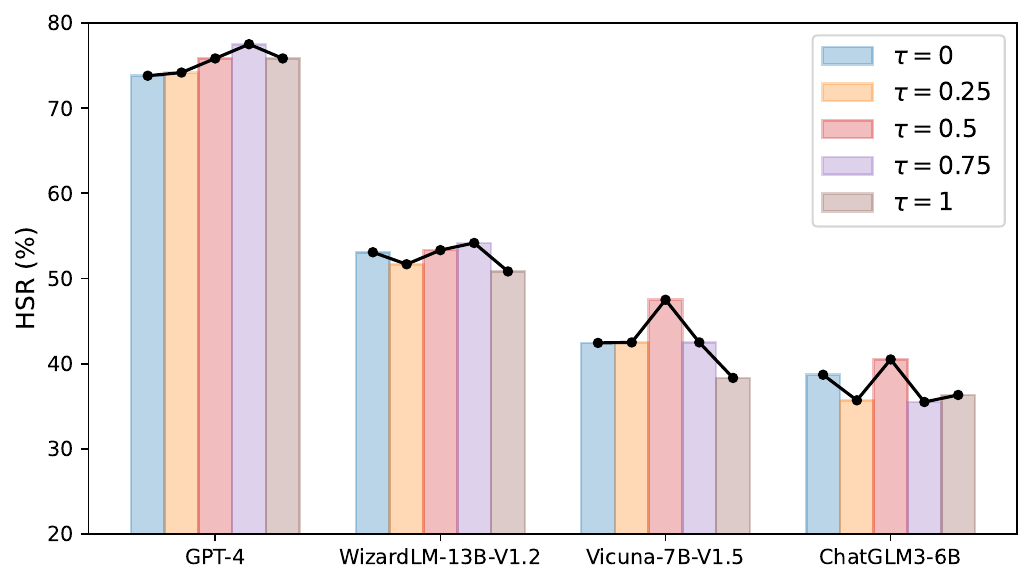}
\caption{
The effect of varying the temperature parameter $\tau$. We use $\tau=0$ to denote greedy decoding.
}
\label{fig:temperature}
\end{figure}

\subsection{Does Different Decoding Strategies Affect the Instruction-following Ability?}
\label{sec: decoding_strategies}
In this section, we systematically investigate the impact of different decoding strategies, represented by the temperature parameter $\tau$, on LLM's instruction-following ability.
The temperature $\tau$ is a commonly used parameter that controls the sharpness of the distribution from which we sample the next token:
\begin{align}
P(w) = \frac{\exp(z_w / \tau)}{\sum_{w' \in V} \exp(z_{w'} / \tau)}
\label{equation: t}
\end{align}
where $z_w$ is the logit for word $w$, $V$ is the vocabulary. Lower values for temperature result in more consistent outputs, while higher values generate more diverse and creative results.
As illustrated in Figure \ref{fig:temperature}, the temperature $\tau$ has a tangible influence on the instruction-following ability across all four models.
The sweet spot seems to be somewhere in the middle where there's enough variability to capture the nuances and intricacies of complex instructions, yet not so much that the model goes off tangent. This balanced behavior ensures that the model remains within the desired context, producing outputs that align closely with the given instructions while also allowing for a slight creative touch when needed.

\section{Conclusion}

In this paper, we introduce \name, a Multi-level Fine-grained Constraints Following Benchmark tailored for gauging the instruction-following capability of LLMs.
\name covers five \emph{fine-grained} constraint categories and over 50 NLP tasks, utilizes a novel \emph{Multi-level} mechanism for precisely estimating the upper limit of instruction-following capability.
Furthermore, we propose an evaluation protocol with three metrics that seamlessly integrate with the multi-level mechanism.
Our extensive tests over 13 popular LLMs reveal a substantial performance advantage for GPT-4 and GPT-3.5 over their counterparts, and there is still significant room for improving the instruction-following ability of current LLMs.

\section*{Limitations}

While our study contributes valuable insights, it is essential to acknowledge several limitations that warrant consideration.

Firstly, our current investigation is confined to single-round interactions, aiming to offer a controlled environment for evaluation. Future research may extend its scope to multi-round conversations to comprehensively assess the instruction-following proficiency of LLMs in more dynamic and extended dialogues~\cite{DBLP:journals/corr/abs-2401-16745}. 

Secondly, the model-based evaluation framework employed in our experiments, while rigorous, relies on prompt engineering, introducing an inherent imperfection. Despite our meticulous selection of high-performing prompts, the potential for further optimization remains, which may impact the reported evaluation metrics.

Lastly, we refrain from proposing specific solutions to address identified weaknesses of LLMs in instruction following. A plausible avenue for future research involves fine-tuning LLMs using our proposed \name as a benchmark, providing a potential roadmap for enhancing instruction adherence.
We defer the exploration of these aspects to subsequent studies, recognizing the need for a comprehensive examination of LLM capabilities across varying interaction complexities.

\section*{Ethics Statement}
Our paper aims to systemically and precisely evaluate the capability of LLMs to follow natural language instructions.
However, it is essential to bear in mind that malicious instructions have the potential to prompt the model to generate harmful or inappropriate outputs. 
Therefore, ensuring safe and responsible practices when assessing the instruction-following capability of LLMs is of paramount importance. 
In \name, each piece of data undergoes a meticulous human review process to identify and eliminate any potentially harmful instructions or offensive content. This rigorous approach underscores our commitment to maintaining a secure and ethical evaluation framework.


\section*{Acknowledgments}
W.\ Wang was supported by HKUST(GZ) Grant G0101000028, CCF-HuaweiDBC202302,  Guangzhou Municipal Science and Technology Project (No.\ 2023A03J0003, 2023A03J0013 and 2024A03J0621).


\bibliography{anthology,custom}

\begin{thebibliography}{58}
\expandafter\ifx\csname natexlab\endcsname\relax\def\natexlab#1{#1}\fi

\bibitem[{arXiv.org submitters(2023)}]{arxivdataset}
arXiv.org submitters. 2023.
\newblock \href {https://doi.org/10.34740/KAGGLE/DSV/6634444} {arxiv dataset}.

\bibitem[{Bach et~al.(2022)Bach, Sanh, Yong, Webson, Raffel, Nayak, Sharma, Kim, Bari, Fevry, Alyafeai, Dey, Santilli, Sun, Ben-david, Xu, Chhablani, Wang, Fries, Al-shaibani, Sharma, Thakker, Almubarak, Tang, Radev, Jiang, and Rush}]{bach-etal-2022-promptsource}
Stephen Bach, Victor Sanh, Zheng~Xin Yong, Albert Webson, Colin Raffel, Nihal~V. Nayak, Abheesht Sharma, Taewoon Kim, M~Saiful Bari, Thibault Fevry, Zaid Alyafeai, Manan Dey, Andrea Santilli, Zhiqing Sun, Srulik Ben-david, Canwen Xu, Gunjan Chhablani, Han Wang, Jason Fries, Maged Al-shaibani, Shanya Sharma, Urmish Thakker, Khalid Almubarak, Xiangru Tang, Dragomir Radev, Mike Tian-jian Jiang, and Alexander Rush. 2022.
\newblock \href {https://doi.org/10.18653/v1/2022.acl-demo.9} {{P}rompt{S}ource: An integrated development environment and repository for natural language prompts}.
\newblock In \emph{Proceedings of the 60th Annual Meeting of the Association for Computational Linguistics: System Demonstrations}, pages 93--104, Dublin, Ireland. Association for Computational Linguistics.

\bibitem[{Bai et~al.(2023)Bai, Bai, Chu, Cui, Dang, Deng, Fan, Ge, Han, Huang, Hui, Ji, Li, Lin, Lin, Liu, Liu, Lu, Lu, Ma, Men, Ren, Ren, Tan, Tan, Tu, Wang, Wang, Wang, Wu, Xu, Xu, Yang, Yang, Yang, Yang, Yao, Yu, Yuan, Yuan, Zhang, Zhang, Zhang, Zhang, Zhou, Zhou, Zhou, and Zhu}]{qwen}
Jinze Bai, Shuai Bai, Yunfei Chu, Zeyu Cui, Kai Dang, Xiaodong Deng, Yang Fan, Wenbin Ge, Yu~Han, Fei Huang, Binyuan Hui, Luo Ji, Mei Li, Junyang Lin, Runji Lin, Dayiheng Liu, Gao Liu, Chengqiang Lu, Keming Lu, Jianxin Ma, Rui Men, Xingzhang Ren, Xuancheng Ren, Chuanqi Tan, Sinan Tan, Jianhong Tu, Peng Wang, Shijie Wang, Wei Wang, Shengguang Wu, Benfeng Xu, Jin Xu, An~Yang, Hao Yang, Jian Yang, Shusheng Yang, Yang Yao, Bowen Yu, Hongyi Yuan, Zheng Yuan, Jianwei Zhang, Xingxuan Zhang, Yichang Zhang, Zhenru Zhang, Chang Zhou, Jingren Zhou, Xiaohuan Zhou, and Tianhang Zhu. 2023.
\newblock Qwen technical report.
\newblock \emph{arXiv preprint arXiv:2309.16609}.

\bibitem[{Bai et~al.(2022)Bai, Jones, Ndousse, Askell, Chen, DasSarma, Drain, Fort, Ganguli, Henighan et~al.}]{bai2022training}
Yuntao Bai, Andy Jones, Kamal Ndousse, Amanda Askell, Anna Chen, Nova DasSarma, Dawn Drain, Stanislav Fort, Deep Ganguli, Tom Henighan, et~al. 2022.
\newblock Training a helpful and harmless assistant with reinforcement learning from human feedback.
\newblock \emph{arXiv preprint arXiv:2204.05862}.

\bibitem[{Baichuan(2023)}]{baichuan2023baichuan2}
Baichuan. 2023.
\newblock \href {https://arxiv.org/abs/2309.10305} {Baichuan 2: Open large-scale language models}.
\newblock \emph{arXiv preprint arXiv:2309.10305}.

\bibitem[{Brown et~al.(2020)Brown, Mann, Ryder, Subbiah, Kaplan, Dhariwal, Neelakantan, Shyam, Sastry, Askell, Agarwal, Herbert-Voss, Krueger, Henighan, Child, Ramesh, Ziegler, Wu, Winter, Hesse, Chen, Sigler, Litwin, Gray, Chess, Clark, Berner, McCandlish, Radford, Sutskever, and Amodei}]{brown2020language}
Tom Brown, Benjamin Mann, Nick Ryder, Melanie Subbiah, Jared~D Kaplan, Prafulla Dhariwal, Arvind Neelakantan, Pranav Shyam, Girish Sastry, Amanda Askell, Sandhini Agarwal, Ariel Herbert-Voss, Gretchen Krueger, Tom Henighan, Rewon Child, Aditya Ramesh, Daniel Ziegler, Jeffrey Wu, Clemens Winter, Chris Hesse, Mark Chen, Eric Sigler, Mateusz Litwin, Scott Gray, Benjamin Chess, Jack Clark, Christopher Berner, Sam McCandlish, Alec Radford, Ilya Sutskever, and Dario Amodei. 2020.
\newblock \href {https://proceedings.neurips.cc/paper_files/paper/2020/file/1457c0d6bfcb4967418bfb8ac142f64a-Paper.pdf} {Language models are few-shot learners}.
\newblock In \emph{Advances in Neural Information Processing Systems}, volume~33, pages 1877--1901. Curran Associates, Inc.

\bibitem[{Cettolo et~al.(2012)Cettolo, Girardi, and Federico}]{cettolo2012wit3}
Mauro Cettolo, Christian Girardi, and Marcello Federico. 2012.
\newblock Wit3: Web inventory of transcribed and translated talks.
\newblock In \emph{Proceedings of the Conference of European Association for Machine Translation (EAMT)}, pages 261--268.

\bibitem[{Chang et~al.(2023)Chang, Wang, Wang, Wu, Zhu, Chen, Yang, Yi, Wang, Wang et~al.}]{chang2023survey}
Yupeng Chang, Xu~Wang, Jindong Wang, Yuan Wu, Kaijie Zhu, Hao Chen, Linyi Yang, Xiaoyuan Yi, Cunxiang Wang, Yidong Wang, et~al. 2023.
\newblock A survey on evaluation of large language models.
\newblock \emph{arXiv preprint arXiv:2307.03109}.

\bibitem[{Chen et~al.(2022)Chen, Li, Chen, and Narasimhan}]{chen2022controllable}
Howard Chen, Huihan Li, Danqi Chen, and Karthik Narasimhan. 2022.
\newblock Controllable text generation with language constraints.
\newblock \emph{arXiv preprint arXiv:2212.10466}.

\bibitem[{Chen et~al.(2021)Chen, Tworek, Jun, Yuan, de~Oliveira~Pinto, Kaplan, Edwards, Burda, Joseph, Brockman, Ray, Puri, Krueger, Petrov, Khlaaf, Sastry, Mishkin, Chan, Gray, Ryder, Pavlov, Power, Kaiser, Bavarian, Winter, Tillet, Such, Cummings, Plappert, Chantzis, Barnes, Herbert-Voss, Guss, Nichol, Paino, Tezak, Tang, Babuschkin, Balaji, Jain, Saunders, Hesse, Carr, Leike, Achiam, Misra, Morikawa, Radford, Knight, Brundage, Murati, Mayer, Welinder, McGrew, Amodei, McCandlish, Sutskever, and Zaremba}]{chen2021codex}
Mark Chen, Jerry Tworek, Heewoo Jun, Qiming Yuan, Henrique~Ponde de~Oliveira~Pinto, Jared Kaplan, Harri Edwards, Yuri Burda, Nicholas Joseph, Greg Brockman, Alex Ray, Raul Puri, Gretchen Krueger, Michael Petrov, Heidy Khlaaf, Girish Sastry, Pamela Mishkin, Brooke Chan, Scott Gray, Nick Ryder, Mikhail Pavlov, Alethea Power, Lukasz Kaiser, Mohammad Bavarian, Clemens Winter, Philippe Tillet, Felipe~Petroski Such, Dave Cummings, Matthias Plappert, Fotios Chantzis, Elizabeth Barnes, Ariel Herbert-Voss, William~Hebgen Guss, Alex Nichol, Alex Paino, Nikolas Tezak, Jie Tang, Igor Babuschkin, Suchir Balaji, Shantanu Jain, William Saunders, Christopher Hesse, Andrew~N. Carr, Jan Leike, Josh Achiam, Vedant Misra, Evan Morikawa, Alec Radford, Matthew Knight, Miles Brundage, Mira Murati, Katie Mayer, Peter Welinder, Bob McGrew, Dario Amodei, Sam McCandlish, Ilya Sutskever, and Wojciech Zaremba. 2021.
\newblock \href {http://arxiv.org/abs/2107.03374} {Evaluating large language models trained on code}.

\bibitem[{Cheng and Li(2022)}]{cheng2022replacing}
Pengyu Cheng and Ruineng Li. 2022.
\newblock Replacing language model for style transfer.
\newblock \emph{arXiv preprint arXiv:2211.07343}.

\bibitem[{Cobbe et~al.(2021)Cobbe, Kosaraju, Bavarian, Chen, Jun, Kaiser, Plappert, Tworek, Hilton, Nakano et~al.}]{cobbe2021training}
Karl Cobbe, Vineet Kosaraju, Mohammad Bavarian, Mark Chen, Heewoo Jun, Lukasz Kaiser, Matthias Plappert, Jerry Tworek, Jacob Hilton, Reiichiro Nakano, et~al. 2021.
\newblock Training verifiers to solve math word problems.
\newblock \emph{arXiv preprint arXiv:2110.14168}.

\bibitem[{Du et~al.(2022)Du, Qian, Liu, Ding, Qiu, Yang, and Tang}]{du2022glm}
Zhengxiao Du, Yujie Qian, Xiao Liu, Ming Ding, Jiezhong Qiu, Zhilin Yang, and Jie Tang. 2022.
\newblock Glm: General language model pretraining with autoregressive blank infilling.
\newblock In \emph{Proceedings of the 60th Annual Meeting of the Association for Computational Linguistics (Volume 1: Long Papers)}, pages 320--335.

\bibitem[{Fan et~al.(2018)Fan, Lewis, and Dauphin}]{fan-etal-2018-hierarchical}
Angela Fan, Mike Lewis, and Yann Dauphin. 2018.
\newblock \href {https://doi.org/10.18653/v1/P18-1082} {Hierarchical neural story generation}.
\newblock In \emph{Proceedings of the 56th Annual Meeting of the Association for Computational Linguistics (Volume 1: Long Papers)}, pages 889--898, Melbourne, Australia. Association for Computational Linguistics.

\bibitem[{Geng et~al.(2023)Geng, Gudibande, Liu, Wallace, Abbeel, Levine, and Song}]{geng2023koala}
Xinyang Geng, Arnav Gudibande, Hao Liu, Eric Wallace, Pieter Abbeel, Sergey Levine, and Dawn Song. 2023.
\newblock Koala: A dialogue model for academic research.
\newblock \emph{Blog post, April}, 1.

\bibitem[{Gilardi et~al.(2023)Gilardi, Alizadeh, and Kubli}]{gilardi2023chatgpt}
Fabrizio Gilardi, Meysam Alizadeh, and Ma{\"e}l Kubli. 2023.
\newblock Chatgpt outperforms crowd-workers for text-annotation tasks.
\newblock \emph{arXiv preprint arXiv:2303.15056}.

\bibitem[{Gliwa et~al.(2019)Gliwa, Mochol, Biesek, and Wawer}]{gliwa-etal-2019-samsum}
Bogdan Gliwa, Iwona Mochol, Maciej Biesek, and Aleksander Wawer. 2019.
\newblock \href {https://doi.org/10.18653/v1/D19-5409} {{SAMS}um corpus: A human-annotated dialogue dataset for abstractive summarization}.
\newblock In \emph{Proceedings of the 2nd Workshop on New Frontiers in Summarization}, pages 70--79, Hong Kong, China. Association for Computational Linguistics.

\bibitem[{Graff et~al.(2003)Graff, Kong, Chen, and Maeda}]{graff2003english}
David Graff, Junbo Kong, Ke~Chen, and Kazuaki Maeda. 2003.
\newblock English gigaword.
\newblock \emph{Linguistic Data Consortium, Philadelphia}, 4(1):34.

\bibitem[{Hendrycks et~al.(2021)Hendrycks, Burns, Basart, Zou, Mazeika, Song, and Steinhardt}]{hendryckstest2021}
Dan Hendrycks, Collin Burns, Steven Basart, Andy Zou, Mantas Mazeika, Dawn Song, and Jacob Steinhardt. 2021.
\newblock Measuring massive multitask language understanding.
\newblock \emph{Proceedings of the International Conference on Learning Representations (ICLR)}.

\bibitem[{Huang et~al.(2023)Huang, Kwak, and An}]{huang2023chatgpt}
F~Huang, H~Kwak, and J~An. 2023.
\newblock Is chatgpt better than human annotators? potential and limitations of chatgpt in explaining implicit hate speech. arxiv.

\bibitem[{Jiang et~al.(2023)Jiang, Chan, Chen, and Wang}]{DBLP:conf/emnlp/JiangCCW23}
Yuxin Jiang, Chunkit Chan, Mingyang Chen, and Wei Wang. 2023.
\newblock \href {https://doi.org/10.18653/V1/2023.EMNLP-MAIN.189} {Lion: Adversarial distillation of proprietary large language models}.
\newblock In \emph{Proceedings of the 2023 Conference on Empirical Methods in Natural Language Processing, {EMNLP} 2023, Singapore, December 6-10, 2023}, pages 3134--3154. Association for Computational Linguistics.

\bibitem[{Jiang et~al.(2024)Jiang, Wang, Wu, Zhong, Zeng, Gao, Li, Jiang, Shang, Tang, Liu, and Wang}]{DBLP:journals/corr/abs-2402-11905}
Yuxin Jiang, Yufei Wang, Chuhan Wu, Wanjun Zhong, Xingshan Zeng, Jiahui Gao, Liangyou Li, Xin Jiang, Lifeng Shang, Ruiming Tang, Qun Liu, and Wei Wang. 2024.
\newblock \href {https://doi.org/10.48550/ARXIV.2402.11905} {Learning to edit: Aligning llms with knowledge editing}.
\newblock \emph{CoRR}, abs/2402.11905.

\bibitem[{Kulal et~al.(2019)Kulal, Pasupat, Chandra, Lee, Padon, Aiken, and Liang}]{kulal2019spoc}
Sumith Kulal, Panupong Pasupat, Kartik Chandra, Mina Lee, Oded Padon, Alex Aiken, and Percy~S Liang. 2019.
\newblock Spoc: Search-based pseudocode to code.
\newblock \emph{Advances in Neural Information Processing Systems}, 32.

\bibitem[{Kwan et~al.(2024)Kwan, Zeng, Jiang, Wang, Li, Shang, Jiang, Liu, and Wong}]{DBLP:journals/corr/abs-2401-16745}
Wai{-}Chung Kwan, Xingshan Zeng, Yuxin Jiang, Yufei Wang, Liangyou Li, Lifeng Shang, Xin Jiang, Qun Liu, and Kam{-}Fai Wong. 2024.
\newblock \href {https://doi.org/10.48550/ARXIV.2401.16745} {Mt-eval: {A} multi-turn capabilities evaluation benchmark for large language models}.
\newblock \emph{CoRR}, abs/2401.16745.

\bibitem[{Li et~al.(2021)Li, Ji, and Han}]{DBLP:conf/naacl/LiJH21}
Sha Li, Heng Ji, and Jiawei Han. 2021.
\newblock \href {https://doi.org/10.18653/v1/2021.naacl-main.69} {Document-level event argument extraction by conditional generation}.
\newblock In \emph{Proceedings of the 2021 Conference of the North American Chapter of the Association for Computational Linguistics: Human Language Technologies, {NAACL-HLT} 2021, Online, June 6-11, 2021}, pages 894--908. Association for Computational Linguistics.

\bibitem[{Li et~al.(2023)Li, Zhang, Dubois, Taori, Gulrajani, Guestrin, Liang, and Hashimoto}]{alpaca_eval}
Xuechen Li, Tianyi Zhang, Yann Dubois, Rohan Taori, Ishaan Gulrajani, Carlos Guestrin, Percy Liang, and Tatsunori~B. Hashimoto. 2023.
\newblock Alpacaeval: An automatic evaluator of instruction-following models.
\newblock \url{https://github.com/tatsu-lab/alpaca_eval}.

\bibitem[{Lison and Tiedemann(2016)}]{lison-tiedemann-2016-opensubtitles2016}
Pierre Lison and J{\"o}rg Tiedemann. 2016.
\newblock \href {https://aclanthology.org/L16-1147} {{O}pen{S}ubtitles2016: Extracting large parallel corpora from movie and {TV} subtitles}.
\newblock In \emph{Proceedings of the Tenth International Conference on Language Resources and Evaluation ({LREC}'16)}, pages 923--929, Portoro{\v{z}}, Slovenia. European Language Resources Association (ELRA).

\bibitem[{Liu et~al.(2023)Liu, Yu, Zhang, Xu, Lei, Lai, Gu, Ding, Men, Yang et~al.}]{liu2023agentbench}
Xiao Liu, Hao Yu, Hanchen Zhang, Yifan Xu, Xuanyu Lei, Hanyu Lai, Yu~Gu, Hangliang Ding, Kaiwen Men, Kejuan Yang, et~al. 2023.
\newblock Agentbench: Evaluating llms as agents.
\newblock \emph{arXiv preprint arXiv:2308.03688}.

\bibitem[{Miller(1992)}]{miller-1992-wordnet}
George~A. Miller. 1992.
\newblock \href {https://aclanthology.org/H92-1116} {{W}ord{N}et: A lexical database for {E}nglish}.
\newblock In \emph{Speech and Natural Language: Proceedings of a Workshop Held at Harriman, New York, {F}ebruary 23-26, 1992}.

\bibitem[{Mishra et~al.(2022)Mishra, Khashabi, Baral, and Hajishirzi}]{mishra-etal-2022-cross}
Swaroop Mishra, Daniel Khashabi, Chitta Baral, and Hannaneh Hajishirzi. 2022.
\newblock \href {https://doi.org/10.18653/v1/2022.acl-long.244} {Cross-task generalization via natural language crowdsourcing instructions}.
\newblock In \emph{Proceedings of the 60th Annual Meeting of the Association for Computational Linguistics (Volume 1: Long Papers)}, pages 3470--3487, Dublin, Ireland. Association for Computational Linguistics.

\bibitem[{Mostafazadeh et~al.(2016)Mostafazadeh, Chambers, He, Parikh, Batra, Vanderwende, Kohli, and Allen}]{mostafazadeh-etal-2016-corpus}
Nasrin Mostafazadeh, Nathanael Chambers, Xiaodong He, Devi Parikh, Dhruv Batra, Lucy Vanderwende, Pushmeet Kohli, and James Allen. 2016.
\newblock \href {https://doi.org/10.18653/v1/N16-1098} {A corpus and cloze evaluation for deeper understanding of commonsense stories}.
\newblock In \emph{Proceedings of the 2016 Conference of the North {A}merican Chapter of the Association for Computational Linguistics: Human Language Technologies}, pages 839--849, San Diego, California. Association for Computational Linguistics.

\bibitem[{Nallapati et~al.(2016)Nallapati, Zhou, Gulcehre, Xiang et~al.}]{nallapati2016abstractive}
Ramesh Nallapati, Bowen Zhou, Caglar Gulcehre, Bing Xiang, et~al. 2016.
\newblock Abstractive text summarization using sequence-to-sequence rnns and beyond.
\newblock \emph{arXiv preprint arXiv:1602.06023}.

\bibitem[{Narayan et~al.(2018)Narayan, Cohen, and Lapata}]{narayan-etal-2018-dont}
Shashi Narayan, Shay~B. Cohen, and Mirella Lapata. 2018.
\newblock \href {https://doi.org/10.18653/v1/D18-1206} {Don{'}t give me the details, just the summary! topic-aware convolutional neural networks for extreme summarization}.
\newblock In \emph{Proceedings of the 2018 Conference on Empirical Methods in Natural Language Processing}, pages 1797--1807, Brussels, Belgium. Association for Computational Linguistics.

\bibitem[{Novikova et~al.(2017)Novikova, Du{\v{s}}ek, and Rieser}]{novikova-etal-2017-e2e}
Jekaterina Novikova, Ond{\v{r}}ej Du{\v{s}}ek, and Verena Rieser. 2017.
\newblock \href {https://doi.org/10.18653/v1/W17-5525} {The {E}2{E} dataset: New challenges for end-to-end generation}.
\newblock In \emph{Proceedings of the 18th Annual {SIG}dial Meeting on Discourse and Dialogue}, pages 201--206, Saarbr{\"u}cken, Germany. Association for Computational Linguistics.

\bibitem[{OpenAI(2023)}]{DBLP:journals/corr/abs-2303-08774}
OpenAI. 2023.
\newblock \href {http://arxiv.org/abs/2303.08774} {{GPT-4} technical report}.
\newblock \emph{CoRR}, abs/2303.08774.

\bibitem[{OpenAI(2022)}]{openai2022chatgpt}
TB~OpenAI. 2022.
\newblock Chatgpt: Optimizing language models for dialogue.
\newblock \emph{OpenAI}.

\bibitem[{Ouyang et~al.(2022)Ouyang, Wu, Jiang, Almeida, Wainwright, Mishkin, Zhang, Agarwal, Slama, Ray et~al.}]{ouyang2022training}
Long Ouyang, Jeffrey Wu, Xu~Jiang, Diogo Almeida, Carroll Wainwright, Pamela Mishkin, Chong Zhang, Sandhini Agarwal, Katarina Slama, Alex Ray, et~al. 2022.
\newblock Training language models to follow instructions with human feedback.
\newblock \emph{Advances in Neural Information Processing Systems}, 35:27730--27744.

\bibitem[{Sanh et~al.(2021)Sanh, Webson, Raffel, Bach, Sutawika, Alyafeai, Chaffin, Stiegler, Scao, Raja et~al.}]{sanh2021multitask}
Victor Sanh, Albert Webson, Colin Raffel, Stephen~H Bach, Lintang Sutawika, Zaid Alyafeai, Antoine Chaffin, Arnaud Stiegler, Teven~Le Scao, Arun Raja, et~al. 2021.
\newblock Multitask prompted training enables zero-shot task generalization.
\newblock \emph{arXiv preprint arXiv:2110.08207}.

\bibitem[{Shanahan et~al.(2023)Shanahan, McDonell, and Reynolds}]{shanahan2023role}
Murray Shanahan, Kyle McDonell, and Laria Reynolds. 2023.
\newblock Role-play with large language models.
\newblock \emph{arXiv preprint arXiv:2305.16367}.

\bibitem[{Smith et~al.(2020)Smith, Gonzalez-Rico, Dinan, and Boureau}]{smith2020controlling}
Eric~Michael Smith, Diana Gonzalez-Rico, Emily Dinan, and Y-Lan Boureau. 2020.
\newblock Controlling style in generated dialogue.
\newblock \emph{arXiv preprint arXiv:2009.10855}.

\bibitem[{Suzgun et~al.(2022)Suzgun, Scales, Sch{\"a}rli, Gehrmann, Tay, Chung, Chowdhery, Le, Chi, Zhou, , and Wei}]{suzgun2022challenging}
Mirac Suzgun, Nathan Scales, Nathanael Sch{\"a}rli, Sebastian Gehrmann, Yi~Tay, Hyung~Won Chung, Aakanksha Chowdhery, Quoc~V Le, Ed~H Chi, Denny Zhou, , and Jason Wei. 2022.
\newblock Challenging big-bench tasks and whether chain-of-thought can solve them.
\newblock \emph{arXiv preprint arXiv:2210.09261}.

\bibitem[{Tang et~al.(2023)Tang, Zong, Zhao, Cohan, and Gerstein}]{tang2023struc}
Xiangru Tang, Yiming Zong, Yilun Zhao, Arman Cohan, and Mark Gerstein. 2023.
\newblock Struc-bench: Are large language models really good at generating complex structured data?
\newblock \emph{arXiv preprint arXiv:2309.08963}.

\bibitem[{Tiedemann(2012)}]{tiedemann2012parallel}
J{\"o}rg Tiedemann. 2012.
\newblock Parallel data, tools and interfaces in opus.
\newblock In \emph{Lrec}, volume 2012, pages 2214--2218. Citeseer.

\bibitem[{Tjong Kim~Sang and De~Meulder(2003)}]{tjong-kim-sang-de-meulder-2003-introduction}
Erik~F. Tjong Kim~Sang and Fien De~Meulder. 2003.
\newblock \href {https://aclanthology.org/W03-0419} {Introduction to the {C}o{NLL}-2003 shared task: Language-independent named entity recognition}.
\newblock In \emph{Proceedings of the Seventh Conference on Natural Language Learning at {HLT}-{NAACL} 2003}, pages 142--147.

\bibitem[{Touvron et~al.(2023)Touvron, Martin, Stone, Albert, Almahairi, Babaei, Bashlykov, Batra, Bhargava, Bhosale, Bikel, Blecher, Ferrer, Chen, Cucurull, Esiobu, Fernandes, Fu, Fu, Fuller, Gao, Goswami, Goyal, Hartshorn, Hosseini, Hou, Inan, Kardas, Kerkez, Khabsa, Kloumann, Korenev, Koura, Lachaux, Lavril, Lee, Liskovich, Lu, Mao, Martinet, Mihaylov, Mishra, Molybog, Nie, Poulton, Reizenstein, Rungta, Saladi, Schelten, Silva, Smith, Subramanian, Tan, Tang, Taylor, Williams, Kuan, Xu, Yan, Zarov, Zhang, Fan, Kambadur, Narang, Rodriguez, Stojnic, Edunov, and Scialom}]{touvron2023llama}
Hugo Touvron, Louis Martin, Kevin Stone, Peter Albert, Amjad Almahairi, Yasmine Babaei, Nikolay Bashlykov, Soumya Batra, Prajjwal Bhargava, Shruti Bhosale, Dan Bikel, Lukas Blecher, Cristian~Canton Ferrer, Moya Chen, Guillem Cucurull, David Esiobu, Jude Fernandes, Jeremy Fu, Wenyin Fu, Brian Fuller, Cynthia Gao, Vedanuj Goswami, Naman Goyal, Anthony Hartshorn, Saghar Hosseini, Rui Hou, Hakan Inan, Marcin Kardas, Viktor Kerkez, Madian Khabsa, Isabel Kloumann, Artem Korenev, Punit~Singh Koura, Marie-Anne Lachaux, Thibaut Lavril, Jenya Lee, Diana Liskovich, Yinghai Lu, Yuning Mao, Xavier Martinet, Todor Mihaylov, Pushkar Mishra, Igor Molybog, Yixin Nie, Andrew Poulton, Jeremy Reizenstein, Rashi Rungta, Kalyan Saladi, Alan Schelten, Ruan Silva, Eric~Michael Smith, Ranjan Subramanian, Xiaoqing~Ellen Tan, Binh Tang, Ross Taylor, Adina Williams, Jian~Xiang Kuan, Puxin Xu, Zheng Yan, Iliyan Zarov, Yuchen Zhang, Angela Fan, Melanie Kambadur, Sharan Narang, Aurelien Rodriguez, Robert Stojnic, Sergey Edunov, and Thomas
  Scialom. 2023.
\newblock \href {http://arxiv.org/abs/2307.09288} {Llama 2: Open foundation and fine-tuned chat models}.

\bibitem[{Tsai et~al.(2021)Tsai, Oraby, Perera, Kao, Du, Narayan-Chen, Chung, and Hakkani-Tur}]{tsai-etal-2021-style}
Alicia Tsai, Shereen Oraby, Vittorio Perera, Jiun-Yu Kao, Yuheng Du, Anjali Narayan-Chen, Tagyoung Chung, and Dilek Hakkani-Tur. 2021.
\newblock \href {https://doi.org/10.18653/v1/2021.nlp4convai-1.21} {Style control for schema-guided natural language generation}.
\newblock In \emph{Proceedings of the 3rd Workshop on Natural Language Processing for Conversational AI}, pages 228--242, Online. Association for Computational Linguistics.

\bibitem[{Vrande{\v{c}}i{\'c} and Kr{\"o}tzsch(2014)}]{vrandevcic2014wikidata}
Denny Vrande{\v{c}}i{\'c} and Markus Kr{\"o}tzsch. 2014.
\newblock Wikidata: a free collaborative knowledgebase.
\newblock \emph{Communications of the ACM}, 57(10):78--85.

\bibitem[{Wang et~al.(2023{\natexlab{a}})Wang, Kordi, Mishra, Liu, Smith, Khashabi, and Hajishirzi}]{wang-etal-2023-self-instruct}
Yizhong Wang, Yeganeh Kordi, Swaroop Mishra, Alisa Liu, Noah~A. Smith, Daniel Khashabi, and Hannaneh Hajishirzi. 2023{\natexlab{a}}.
\newblock \href {https://doi.org/10.18653/v1/2023.acl-long.754} {Self-instruct: Aligning language models with self-generated instructions}.
\newblock In \emph{Proceedings of the 61st Annual Meeting of the Association for Computational Linguistics (Volume 1: Long Papers)}, pages 13484--13508, Toronto, Canada. Association for Computational Linguistics.

\bibitem[{Wang et~al.(2023{\natexlab{b}})Wang, Zhong, Li, Mi, Zeng, Huang, Shang, Jiang, and Liu}]{wang2023aligning}
Yufei Wang, Wanjun Zhong, Liangyou Li, Fei Mi, Xingshan Zeng, Wenyong Huang, Lifeng Shang, Xin Jiang, and Qun Liu. 2023{\natexlab{b}}.
\newblock Aligning large language models with human: A survey.
\newblock \emph{arXiv preprint arXiv:2307.12966}.

\bibitem[{Wang et~al.(2023{\natexlab{c}})Wang, Zhang, Yang, Shi, Zhou, Hao, Xiong, Li, Sim, Chen et~al.}]{wang2023interactive}
Zekun Wang, Ge~Zhang, Kexin Yang, Ning Shi, Wangchunshu Zhou, Shaochun Hao, Guangzheng Xiong, Yizhi Li, Mong~Yuan Sim, Xiuying Chen, et~al. 2023{\natexlab{c}}.
\newblock Interactive natural language processing.
\newblock \emph{arXiv preprint arXiv:2305.13246}.

\bibitem[{Weller et~al.(2020)Weller, Lourie, Gardner, and Peters}]{weller-etal-2020-learning}
Orion Weller, Nicholas Lourie, Matt Gardner, and Matthew~E. Peters. 2020.
\newblock \href {https://doi.org/10.18653/v1/2020.emnlp-main.105} {Learning from task descriptions}.
\newblock In \emph{Proceedings of the 2020 Conference on Empirical Methods in Natural Language Processing (EMNLP)}, pages 1361--1375, Online. Association for Computational Linguistics.

\bibitem[{Xu et~al.(2023)Xu, Sun, Zheng, Geng, Zhao, Feng, Tao, and Jiang}]{xu2023wizardlm}
Can Xu, Qingfeng Sun, Kai Zheng, Xiubo Geng, Pu~Zhao, Jiazhan Feng, Chongyang Tao, and Daxin Jiang. 2023.
\newblock \href {http://arxiv.org/abs/2304.12244} {Wizardlm: Empowering large language models to follow complex instructions}.

\bibitem[{Yang et~al.(2023)Yang, Nachum, Du, Wei, Abbeel, and Schuurmans}]{yang2023foundation}
Sherry Yang, Ofir Nachum, Yilun Du, Jason Wei, Pieter Abbeel, and Dale Schuurmans. 2023.
\newblock Foundation models for decision making: Problems, methods, and opportunities.
\newblock \emph{arXiv preprint arXiv:2303.04129}.

\bibitem[{Yao et~al.(2022)Yao, Zhao, Yu, Du, Shafran, Narasimhan, and Cao}]{yao2022react}
Shunyu Yao, Jeffrey Zhao, Dian Yu, Nan Du, Izhak Shafran, Karthik Narasimhan, and Yuan Cao. 2022.
\newblock React: Synergizing reasoning and acting in language models.
\newblock \emph{arXiv preprint arXiv:2210.03629}.

\bibitem[{Zhang et~al.(2022)Zhang, Song, Li, Zhou, and Song}]{zhang2022survey}
Hanqing Zhang, Haolin Song, Shaoyu Li, Ming Zhou, and Dawei Song. 2022.
\newblock A survey of controllable text generation using transformer-based pre-trained language models.
\newblock \emph{ACM Computing Surveys}.

\bibitem[{Zhao et~al.(2023)Zhao, Zhang, Si, Nan, Tang, and Cohan}]{zhao2023large}
Yilun Zhao, Haowei Zhang, Shengyun Si, Linyong Nan, Xiangru Tang, and Arman Cohan. 2023.
\newblock Large language models are effective table-to-text generators, evaluators, and feedback providers.
\newblock \emph{arXiv preprint arXiv:2305.14987}.

\bibitem[{Zheng et~al.(2023)Zheng, Chiang, Sheng, Zhuang, Wu, Zhuang, Lin, Li, Li, Xing, Zhang, Gonzalez, and Stoica}]{vicuna2023}
Lianmin Zheng, Wei-Lin Chiang, Ying Sheng, Siyuan Zhuang, Zhanghao Wu, Yonghao Zhuang, Zi~Lin, Zhuohan Li, Dacheng Li, Eric.~P Xing, Hao Zhang, Joseph~E. Gonzalez, and Ion Stoica. 2023.
\newblock \href {http://arxiv.org/abs/2306.05685} {Judging llm-as-a-judge with mt-bench and chatbot arena}.

\bibitem[{Zhong et~al.(2023)Zhong, Cui, Guo, Liang, Lu, Wang, Saied, Chen, and Duan}]{zhong2023agieval}
Wanjun Zhong, Ruixiang Cui, Yiduo Guo, Yaobo Liang, Shuai Lu, Yanlin Wang, Amin Saied, Weizhu Chen, and Nan Duan. 2023.
\newblock \href {https://doi.org/10.48550/arXiv.2304.06364} {Agieval: {A} human-centric benchmark for evaluating foundation models}.
\newblock \emph{CoRR}, abs/2304.06364.

\end{thebibliography}

\appendix

\section{Data Generation Process}
\label{appendix: data_generation_process}
Here we outline the sources for our data and provide a detailed description of the data generation process for each constraint category.

\subsection{Content Constraints}
\label{appendix: data_generation_content}
\begin{figure*}[!h]
\centering
\includegraphics[width=\linewidth]{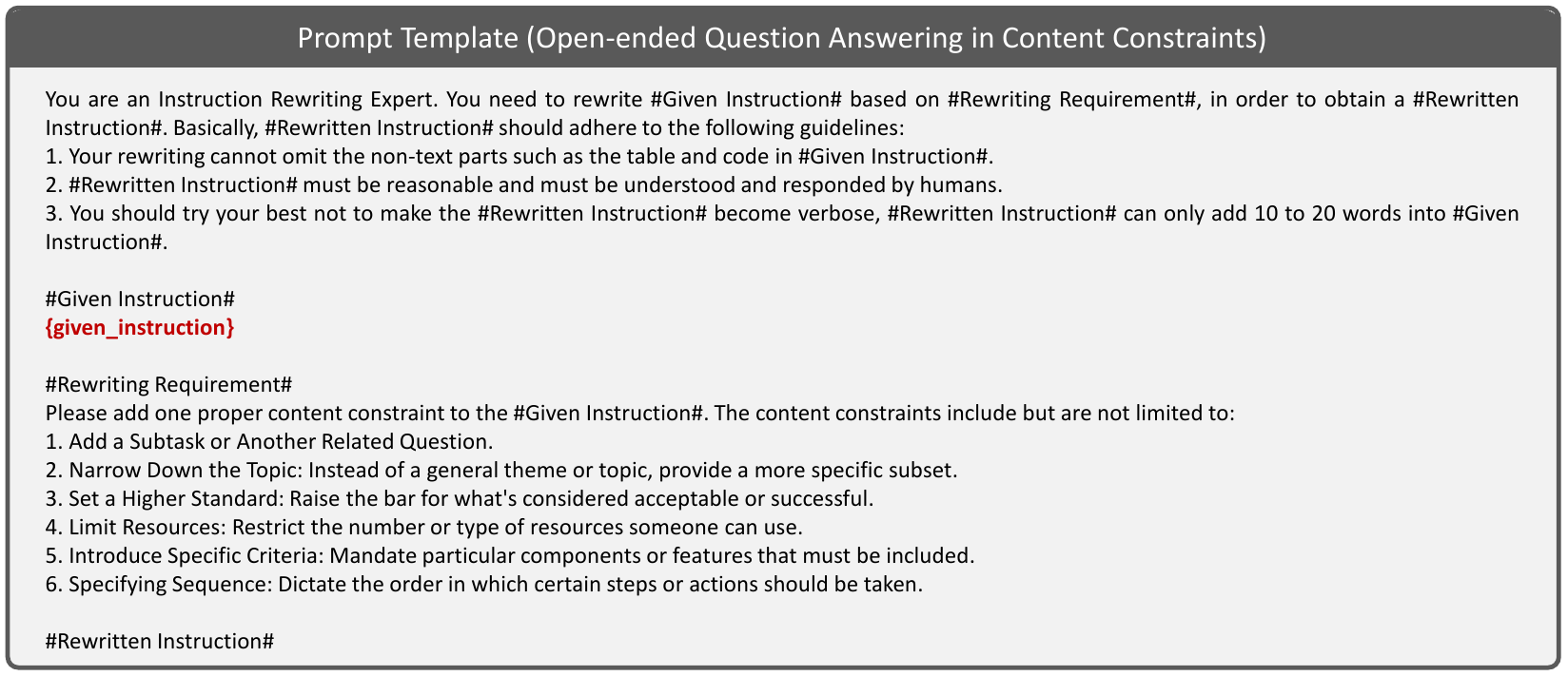}
\caption{
The prompt template for Open-ended Question Answering in Content Constraints.
}
\label{fig:content_prompt}
\end{figure*}

The data of content constraints is constructed from five tasks as follows:

\begin{itemize}
    \item \textbf{Data-to-Text Generation}
    We create instructions with 1 to 5 constraints by adapting samples from E2E~\cite{novikova-etal-2017-e2e}. Different from the original task, we ask the model to extract the flat meaning representations according to the corresponding natural language texts. The number of constraints increases with the number of attributes and the number of restaurants. We use exact match as the evaluation metric.
    
    \item \textbf{Document-Level Event Argument Extraction}
    We create instructions by adapting samples from WIKIEVENTS~\cite{DBLP:conf/naacl/LiJH21}. Given a document, the model is required to extract $n$ events that satisfy a specific event template, where $n\in[1, 5]$ corresponds to the number of constraints. We use accuracy as the evaluation metric.
    
    \item \textbf{Document-Level Named Entity Recognition}
    We derive instructions from samples in the CONLL-2003 dataset~\cite{tjong-kim-sang-de-meulder-2003-introduction}. We ask the model to extract a single named entity from a provided document. Notably, as the number of constraints rises, the requirements for the retrieved named entity correspondingly increase. For example, ``extract one named entity that is a location'' $\rightarrow$ ``extract one named entity that is a location in east Asia''. We use accuracy as the evaluation metric.

    \item \textbf{Text Generation with Language Constraints} 
    COGNAC~\cite{chen2022controllable} is a challenging benchmark wherein models are presented with a topic accompanied by example text and explicit constraints on the text to avoid. We curate data from COGNAC, formulating instructions with 1 to 5 constraints by integrating additional linguistic restrictions from WordNet~\cite{miller-1992-wordnet} and Wikidata~\cite{vrandevcic2014wikidata}.

    \item \textbf{Open-ended Question Answering}
    We first choose initial instructions from existing datasets including self-instruct evaluation set~\cite{wang-etal-2023-self-instruct}, helpful evaluation released by Anthropic\cite{bai2022training}, Vicuna evaluation\cite{vicuna2023}, and Koala evaluation\cite{geng2023koala}, as well as opensource platforms such as Quora \footnote{\url{https://www.quora.com}}, Reddit \footnote{\url{https://www.reddit.com}}, and ShareGPT \footnote{\url{https://sharegpt.com}}. Given the challenges associated with iteratively adding constraints to an initial instruction, we prompt GPT-4 with a specific prompt shown in Figure \ref{fig:content_prompt} to generate a new instruction with one more constraint based on the given instruction. 
    The above process is repeated five times. Finally, we obtain a set of instructions ranging from 1 to 5 constraints. 
\end{itemize}

\subsection{Situation Constraints}
\label{appendix: data_generation_scenario}
The data of situation constraints is constructed from tasks as follows:
\begin{itemize}

    \item \textbf{Suggestion Generation, Role-playing}
    We collect multi-level instructions that fit within the paradigm of situation constraints from Open-ended Question Answering datasets and online platforms. Examples include asking the model to give suggestions under specific circumstances, asking the model to act as a terminal and output based on the given information, etc.
    
    \item \textbf{Math Word Problems}
    The initial instructions are collected from GSM8K~\cite{cobbe2021training} and AGIEval~\cite{zhong2023agieval}. We then manually add constraints progressively by enhancing the situation descriptions, ensuring that the core question remains unaltered. We use accuracy as the evaluation metric.

    \item \textbf{Time/Spatial Reasoning}
    We generate data by refining samples from BIG-Bench Hard~\cite{suzgun2022challenging}. For Time Reasoning, we increase the difficulty level by incorporating additional temporal concepts, such as weeks, months, and years. In the realm of Spatial Reasoning, we opt for a logical deduction task that necessitates deducing the order of a sequence of objects. Here, the number of constraints escalates by augmenting the task with detailed location descriptions for a new object. We use accuracy as the evaluation metric.
    
    \item \textbf{Code Generation}
    We sourced initial instructions from HumanEval~\cite{chen2021codex} and enhanced the difficulty level by adding complexity to the function descriptions within the instructions. We use pass@1~\cite{kulal2019spoc} as the evaluation metric.

\end{itemize}

\subsection{Example Constraints}
\label{appendix: data_generation_example}
Specifically, we choose 40 diverse NLP tasks from PromptSource~\cite{bach-etal-2022-promptsource}, where each task has more than 5 question templates.
Additionally, we create 29 answer templates (shown in Table \ref{tab:answer_template}) that regulate the format of the response.
For instructions at difficulty level 1, we utilize the standard 5-shot prompting, where 5 shots are equipped with 1 sampled question template and 1 sampled answer template, and the model is required to respond to a query using the answer template. For instructions at difficulty level $n\ (1<n\le 5)$, the 5 shots are randomly paired with $n$ question templates and $n$ corresponding answer templates. Based on the question template of the query, the model is required to recognize the matched question template in the 5 shots and respond using the corresponding answer template. We use accuracy as the evaluation metric.

\begin{table}[t]
\small
\centering
\begin{tabular}{l}
\toprule
\textbf{Answer template} \\
\midrule
\{question\}\textbackslash n\{answer\} \\
\{question\}\textbackslash nA: \{answer\} \\
\{question\}\textbackslash nAnswer: \{answer\} \\
\{question\}\textbackslash nANSWER: \{answer\} \\
\{question\}\textbackslash n[Answer]\textbackslash n\{answer\} \\
\{question\}\textbackslash n\#Answer\#\textbackslash n\{answer\} \\
\{question\}\textbackslash nThe answer is: \{answer\} \\
\{question\}\textbackslash n\{"answer": "\{answer\}"\} \\
\{question\}\textbackslash n\{"Answer": "\{answer\}"\} \\
\{question\}\textbackslash n<body>\{answer\}</body> \\
\{question\}\textbackslash nResponse: \{answer\} \\
\{question\}\textbackslash nRESPONSE: \{answer\} \\
\{question\}\textbackslash n[Response]\textbackslash n\{answer\} \\
\{question\}\textbackslash n\#Response\#\textbackslash n\{answer\} \\
\{question\}\textbackslash nThe response is: \{answer\} \\
\{question\}\textbackslash n\{"response": "\{answer\}"\} \\
\{question\}\textbackslash n\{"Response": "\{answer\}"\} \\
\{question\}\textbackslash nBot: \{answer\} \\
\{question\}\textbackslash nBOT: \{answer\} \\
\{question\}\textbackslash n[Bot]\textbackslash n\{answer\} \\
\{question\}\textbackslash n\#Bot\#\textbackslash n\{answer\} \\
\{question\}\textbackslash nThe response of the bot is: \{answer\} \\
\{question\}\textbackslash n\{"bot": "\{answer\}"\} \\
\{question\}\textbackslash n\{"Bot": "\{answer\}"\} \\
\{question\}\textbackslash nAI assistant: \{answer\} \\
\{question\}\textbackslash n[AI assistant]\textbackslash n\{answer\} \\
\{question\}\textbackslash n\#AI assistant\#\textbackslash n\{answer\} \\
\{question\}\textbackslash nThe response of the AI assistant is: \{answer\} \\
\{question\}\textbackslash n\{"AI assistant": "\{answer\}"\} \\
\bottomrule
\end{tabular}
\caption{Answer template of Example Constraints.}
\label{tab:answer_template}
\end{table}

\subsection{Mixed Constraints}
\label{appendix: data_generation_mixed}
In this paper, we consider four below tasks which are naturally suitable for constructing mixed constraints:

\begin{itemize}

    \item \textbf{Text Editing} 
    We start by gathering text from different online sources, like sentences, letters, and emails. Next, we create instructions with multi-level mixed constraints by increasingly adding an editing requirement to the text at each level. For example, ``swap the first and last words in the sentence'' (Content Constraints), ``response using '\#\#\#' at the beginning'' (Format Constraints), etc.
    We write rule-based programs for individual instructions to assess the satisfaction of internal constraints, employing exact match as the evaluation metric.
    
    \item \textbf{Summarization} The initial instructions are sampled from CNN/Daily Mail\cite{nallapati2016abstractive},
    XSum~\cite{narayan-etal-2018-dont}, SAMSum~\cite{gliwa-etal-2019-samsum}, English Gigaword~\cite{graff2003english}, and arXiv~\cite{arxivdataset}.
    The instructions with multi-level mixed constraints are produced by specifying the format of generating answers (Format Constraints), requiring the generated text to include or not include certain keywords (Content Constraints), etc.
    We write rule-based programs for individual instructions to assess the satisfaction of internal constraints, employing accuracy as the evaluation metric.

    \item \textbf{Machine Translation} The initial instructions are sampled from OpenSubtitles~\cite{lison-tiedemann-2016-opensubtitles2016}, TED Talks~\cite{cettolo2012wit3}, and News-Commentary~\cite{tiedemann2012parallel}.
    Then we construct instructions from level 1 to level 5 using a similar pipeline as that of Summarization.
    We write rule-based programs for individual instructions to assess the satisfaction of internal constraints, employing accuracy as the evaluation metric.

    \item \textbf{Story Generation} We collect initial instructions from ROCStories~\cite{mostafazadeh-etal-2016-corpus} and WritingPrompts~\cite{fan-etal-2018-hierarchical}. Then we add 5 mixed constraints sequentially to the initial instructions based on the ground truth, such as the number of sentences in the generated story (Format Constraints), requiring the generated text to include certain keywords (Content Constraints), specifying the writing style (Style Constraints), etc.

\end{itemize}



\begin{figure*}[!h]
\centering
\includegraphics[width=\linewidth]{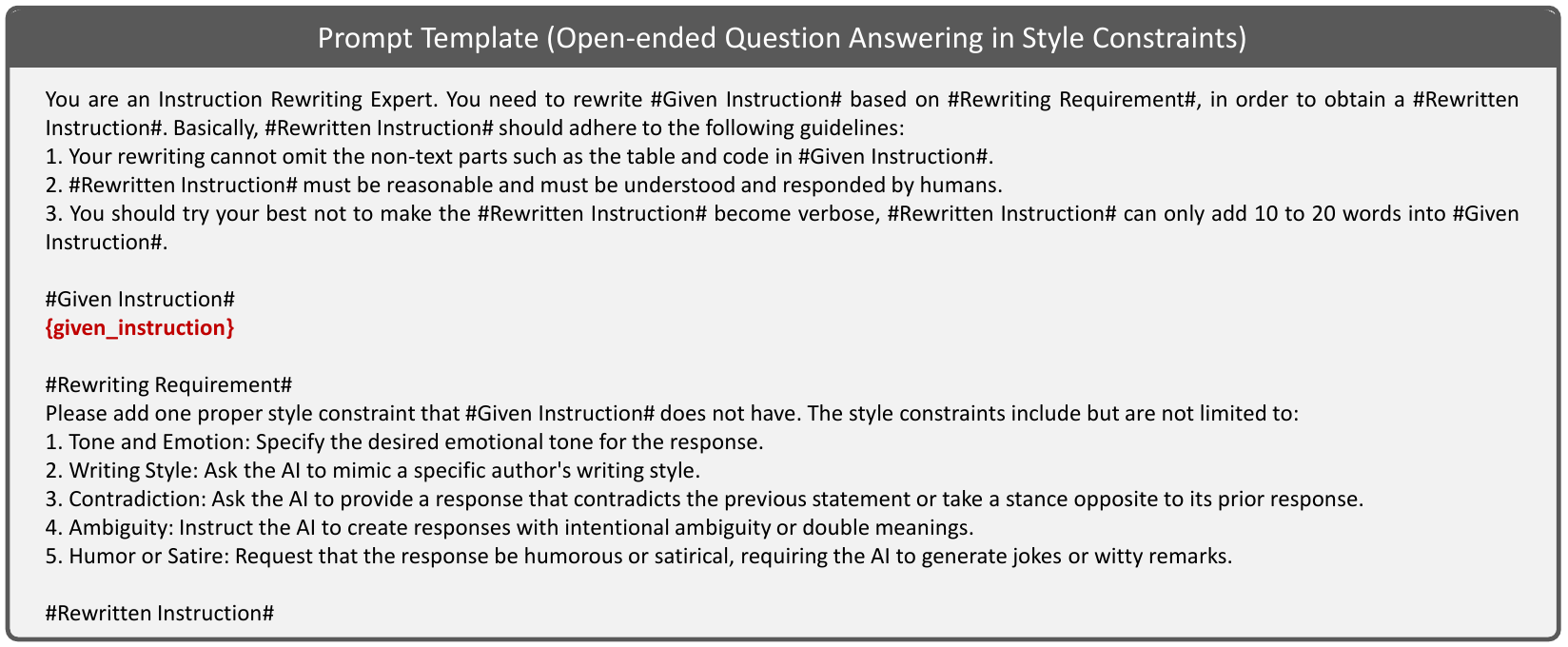}
\caption{
The prompt template for Open-ended Question Answering in Style Constraints.
}
\label{fig:style_prompt}
\end{figure*}

\begin{figure*}[!h]
\centering
\includegraphics[width=\linewidth]{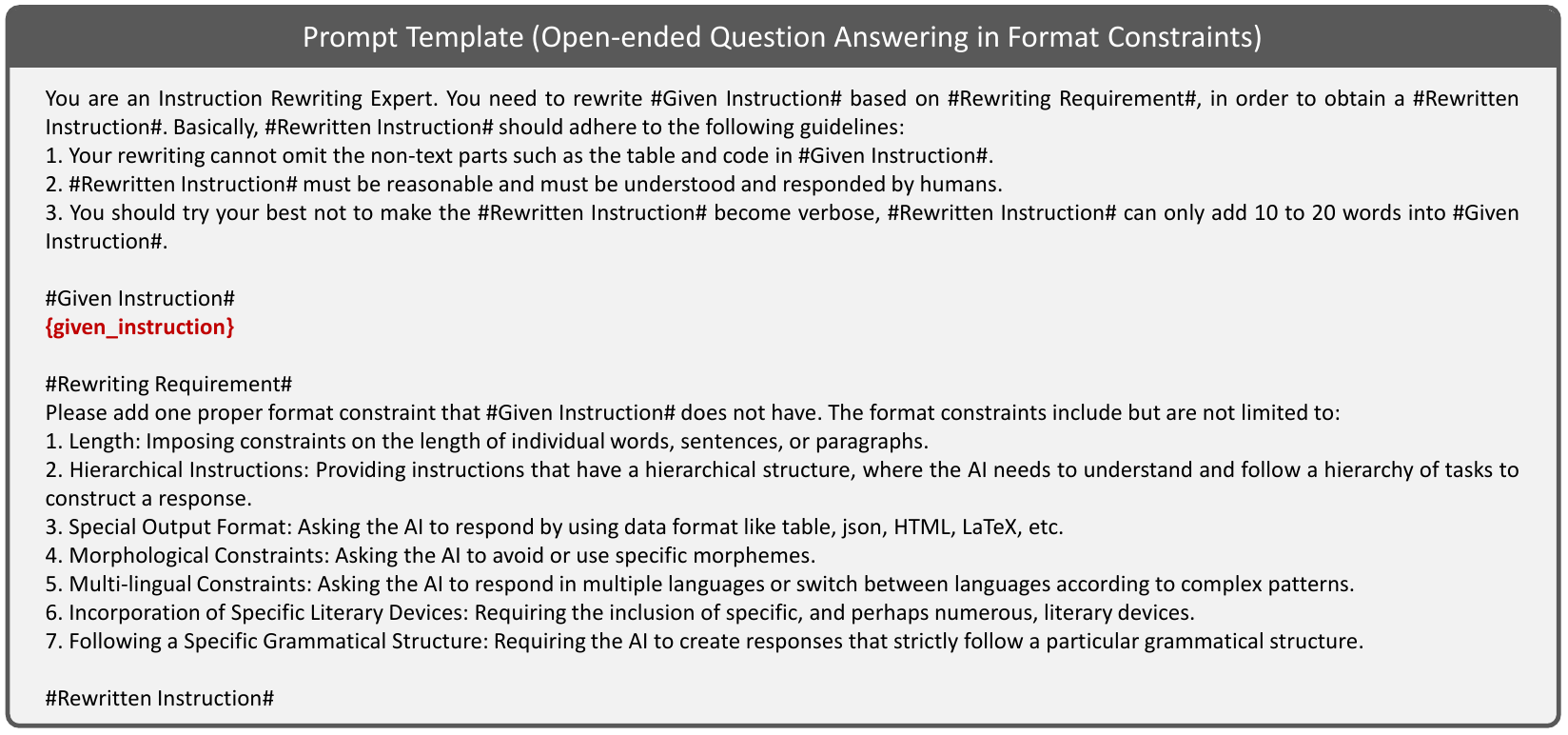}
\caption{
The prompt template for Open-ended Question Answering in Format Constraints.
}
\label{fig:format_prompt}
\end{figure*}

\section{Detailed Experimental Results}
\label{appendix: experiments}
Here we list the experimental results across 5 difficulty levels for each constraint category, including Content Constraints in Table \ref{tab:content_level}, Situation Constraints in Table \ref{tab:scenario_level}, Style Constraints in Table \ref{tab:style_level}, Format Constraints in Table \ref{tab:format_level}, Example Constraints in Table \ref{tab:example_level}, and Mixed Constraints in Table \ref{tab:mixed_level}.

\begin{table*}[!h]
\small
\centering
\setlength{\tabcolsep}{6pt}
\begin{tabular}{l|cccccc|cccccc|c}
\toprule
& \multicolumn{6}{c|}{\textbf{HSR (\%)}} & \multicolumn{6}{c|}{\textbf{SSR (\%)}} \\
\cmidrule(lr){2-7} \cmidrule(lr){8-13}
\multirow{-2}{*}{\centering\textbf{Model}}
& \textbf{L1} & \textbf{L2} & \textbf{L3} & \textbf{L4} & \textbf{L5} & \textbf{Avg.} & \textbf{L1} & \textbf{L2} & \textbf{L3} & \textbf{L4} & \textbf{L5} & \textbf{Avg.}
& \multirow{-2}{*}{\centering\textbf{CSL}} \\
\midrule
\rowcolor{colorGPT}
GPT-4-Preview-1106 & \textbf{84.0} & \textbf{76.0} & \textbf{72.0} & \textbf{80.0} & \textbf{72.0} & \textbf{76.8} & \textbf{84.0} &\textbf{78.0} & 74.7 & \textbf{83.0} & \textbf{80.8} & \textbf{80.1} & \textbf{3.5} \\
\rowcolor{colorGPT}
GPT-3.5-Turbo-1106 & 72.0 & 68.0 & \textbf{72.0} & 56.0 & 48.0 & 63.2 & 72.0 & 70.0 & \textbf{76.0} & 67.0 & 64.0 & 69.8 & 2.7 \\
\rowcolor{color70B}
Qwen-Chat-72B & \textbf{84.0} & 72.0 & \textbf{72.0} & 52.0 & 56.0 & 67.2 & \textbf{84.0} & 76.0 & \textbf{76.0} & 62.0 & 66.4 & 72.9 & 3.2 
\\
\rowcolor{color70B}
LLaMA2-Chat-70B & 48.0	&44.0	&44.0	&40.0	&40.0 &43.2 &48.0	&48.0	&48.0	&47.0	&47.2  &47.6 &2.2
\\
\rowcolor{color13B}
Qwen-Chat-14B & 56.0 & 64.0 & 60.0 & 44.0 & 36.0 &52.0 & 56.0 & 68.0 & 66.7 & 55.0 & 51.2 &59.4 & 2.1 \\
\rowcolor{color13B}
WizardLM-13B-V1.2 & 68.0 & 56.0 & 48.0 & 44.0 & 28.0 &48.8 & 68.0 & 60.0 & 56.0 & 51.0 & 45.6 & 56.1 & 2.4 \\
\rowcolor{color13B}
LLaMA2-Chat-13B & 48.0 & 44.0 & 48.0 & 48.0 & 36.0 &44.8 & 48.0 & 48.0 & 50.7 & 50.0 & 47.2 &48.8 & 2.1 \\
\rowcolor{color13B}
Vicuna-13B-V1.5 & 60.0 & 52.0 & 52.0 & 44.0 & 32.0 &48.0 & 60.0 & 58.0 & 58.7 & 53.0 & 44.8 &54.9 & 2.3 \\
\rowcolor{colorOther}
Qwen-Chat-7B & 56.0 & 56.0 & 36.0 & 36.0 & 24.0 &41.6 & 56.0 & 60.0 & 48.0 & 48.0 & 40.8 & 50.6 & 1.7 \\
\rowcolor{colorOther}
LLaMA2-Chat-7B & 44.0 & 48.0 & 44.0 & 40.0 & 36.0 &42.4 & 44.0 & 48.0 & 46.7 & 46.0 & 46.4 &46.2 & 1.8 \\
\rowcolor{colorOther}
Vicuna-7B-V1.5 & 60.0 & 48.0 & 52.0 & 40.0 & 16.0 &43.2 & 60.0 & 56.0 & 61.3 & 51.0 & 44.0 &54.5 & 1.9 \\
\rowcolor{colorOther}
Baichuan2-Chat-7B & 60.0 & 48.0 & 40.0 & 36.0 & 24.0 &41.6 & 60.0 & 52.0 & 45.3 & 50.0 & 44.8 &50.4 & 1.7 \\
\rowcolor{colorOther}
ChatGLM3-6B  & 68.0&	44.0&	44.0&	36.0&	24.0 &43.2 &	68.0&	52.0&	49.3&	50.0&	40.8 &52.0 &	1.9 \\
\bottomrule
\end{tabular}
\caption{Results of Content Constraints across 5 difficulty levels. \colorbox{colorGPT}{Proprietary LLMs}, \colorbox{color70B}{open-sourced LLMs (large)}, \colorbox{color13B}{open-sourced LLMs (medium)}, and \colorbox{colorOther}{open-sourced LLMs (small)} are distinguished by different colors.}
\label{tab:content_level}
\end{table*}

\begin{table*}[!h]
\small
\centering
\setlength{\tabcolsep}{6pt}
\begin{tabular}{l|cccccc|cccccc|c}
\toprule
& \multicolumn{6}{c|}{\textbf{HSR (\%)}} & \multicolumn{6}{c|}{\textbf{SSR (\%)}} \\
\cmidrule(lr){2-7} \cmidrule(lr){8-13}
\multirow{-2}{*}{\centering\textbf{Model}}
& \textbf{L1} & \textbf{L2} & \textbf{L3} & \textbf{L4} & \textbf{L5} & \textbf{Avg.} & \textbf{L1} & \textbf{L2} & \textbf{L3} & \textbf{L4} & \textbf{L5} & \textbf{Avg.}
& \multirow{-2}{*}{\centering\textbf{CSL}} \\
\midrule
\rowcolor{colorGPT}
GPT-4-Preview-1106 & \textbf{90.0} & \textbf{90.0} & \textbf{85.0} & \textbf{65.0} & 50.0 &\textbf{76.0} & \textbf{90.0} & \textbf{90.0} & \textbf{88.3} & \textbf{76.2} & 69.0 &\textbf{82.7} & \textbf{3.5} \\
\rowcolor{colorGPT}
GPT-3.5-Turbo-1106 & 72.7 & 72.7 & 72.7 & 63.6 & \textbf{68.2} & 70.0 & 72.7 & 72.7 & 75.8 & 71.6 & \textbf{75.5} &73.7 & 3.2 \\
\rowcolor{color70B}
Qwen-Chat-72B & 81.8	 &81.8	 &54.6	 &54.6	 &50.0  &64.6 &81.8 &86.4	&71.2	&65.9	&60.9 &73.2 &2.4 \\
\rowcolor{color70B}
LLaMA2-Chat-70B & 72.7	 &68.2	 &54.6	 &40.9	 &50.0  &57.3 &72.7	&70.5	&66.7	&61.4	&68.2 &67.9 &2.4 \\
\rowcolor{color13B}
Qwen-Chat-14B & 72.7	 &68.2	 &59.1	 &45.5	 &50.0  &59.1 &72.7	&72.7	&71.2	&56.8	&61.8 &67.1 &2.2 \\
\rowcolor{color13B}
WizardLM-13B-V1.2 & 65.0 & 65.0 & 70.0 & 35.0 & 45.0 &56.0 & 65.0 & 67.5 & 71.7 & 50.0 & 58.0 &62.4 & 1.9 \\
\rowcolor{color13B}
LLaMA2-Chat-13B & 63.6 & 77.3 & 59.1 & 45.5 & 36.4 &56.4 & 63.6 & 81.8 & 69.7 & 58.0 & 53.6 &65.4 & 2.2 \\
\rowcolor{color13B}
Vicuna-13B-V1.5 & 68.2 & 63.6 & 54.5 & 31.8 & 40.9 &51.8 & 68.2 & 65.9 & 60.6 & 47.7 & 54.5 &59.4 & 1.9 \\
\rowcolor{colorOther}
Qwen-Chat-7B & 59.1	& 59.1 & 54.6 &	36.4 &	50.0 &	51.8 &  59.1 & 68.2 & 65.2 & 52.3 & 66.4 & 62.2  & 1.6 \\
\rowcolor{colorOther}
LLaMA2-Chat-7B & 68.2 & 45.5 & 54.5 & 27.3 & 54.5 &50.0 & 68.2 & 59.1 & 63.6 & 54.5 & 65.5  &62.2 & 1.8 \\
\rowcolor{colorOther}
Vicuna-7B-V1.5 & 45.5 & 45.5 & 31.8 & 22.7 & 27.3 &34.5 & 45.5 & 50.0 & 43.9 & 34.1 & 49.1 &44.5 & 1.4 \\
\rowcolor{colorOther}
Baichuan2-Chat-7B & 36.4 & 40.9 & 40.9 & 22.7 & 18.2 &31.8 & 36.4 & 54.5 & 54.5 & 42.0 & 41.8 &45.9 & 0.9 \\
\rowcolor{colorOther}
ChatGLM3-6B  & 63.6&	63.6&	40.9&	27.3&	22.7&	43.6 & 63.6&	70.5&	56.1&	44.3&	43.6&	55.6 & 1.8
   \\
\bottomrule
\end{tabular}
\caption{Results of Situation Constraints across 5 difficulty levels. \colorbox{colorGPT}{Proprietary LLMs}, \colorbox{color70B}{open-sourced LLMs (large)}, \colorbox{color13B}{open-sourced LLMs (medium)}, and \colorbox{colorOther}{open-sourced LLMs (small)} are distinguished by different colors.}
\label{tab:scenario_level}
\end{table*}

\begin{table*}[!h]
\small
\centering
\setlength{\tabcolsep}{6pt}
\begin{tabular}{l|cccccc|cccccc|c}
\toprule
& \multicolumn{6}{c|}{\textbf{HSR (\%)}} & \multicolumn{6}{c|}{\textbf{SSR (\%)}} \\
\cmidrule(lr){2-7} \cmidrule(lr){8-13}
\multirow{-2}{*}{\centering\textbf{Model}}
& \textbf{L1} & \textbf{L2} & \textbf{L3} & \textbf{L4} & \textbf{L5} & \textbf{Avg.} & \textbf{L1} & \textbf{L2} & \textbf{L3} & \textbf{L4} & \textbf{L5} & \textbf{Avg.}
& \multirow{-2}{*}{\centering\textbf{CSL}} \\
\midrule
\rowcolor{colorGPT}
GPT-4-Preview-1106 & \textbf{96.7} & \textbf{93.3} & 86.7 & \textbf{96.7} & \textbf{90.0} & \textbf{92.7} & \textbf{96.7} & 95.0 & 93.3 & \textbf{98.3} & \textbf{98.0} & 96.3 & \textbf{4.3} \\
\rowcolor{colorGPT}
GPT-3.5-Turbo-1106 & \textbf{96.7} & \textbf{93.3} & 90.0 & 93.3 & 86.7 & 92.0 & \textbf{96.7} & \textbf{96.7} & 95.6 & \textbf{98.3} & 97.3 &96.9 & 4.1 \\
\rowcolor{color70B}
Qwen-Chat-72B & 90.0 &83.3 &70.0 &66.7 &56.7 &73.3 &90.0 &90.0 &90.0 &89.2 &86.0 & 89.0 & 3.0 \\
\rowcolor{color70B}
LLaMA2-Chat-70B & \textbf{96.7}	&\textbf{93.3}	&\textbf{93.3}	&83.3	&83.3 &90.0 &\textbf{96.7}	&\textbf{96.7}	&\textbf{97.8}	&95.0	&96.0 &\textbf{96.4} &4.1 \\
\rowcolor{color13B}
Qwen-Chat-14B  & 80.0 & 73.3 & 46.7 &60.0 & 46.7 & 61.3 & 80.0 & 86.7 & 75.6 & 82.5 & 78.0 &80.6 & 2.2 \\
\rowcolor{color13B}
WizardLM-13B-V1.2 & \textbf{96.7} & \textbf{93.3} & 80.0 & 83.3 & 60.0 &82.7 & \textbf{96.7} & 95.0 & 91.1 & 92.5 & 90.0 &93.1 & 3.6 \\
\rowcolor{color13B}
LLaMA2-Chat-13B & \textbf{96.7} & \textbf{93.3} & 90.0 & 86.7 & 86.7 &90.7 & \textbf{96.7} & \textbf{96.7} & 95.6 & 96.7 & 96.0 &96.3 & 4.1 \\
\rowcolor{color13B}
Vicuna-13B-V1.5 & 90.0 & 90.0 & 60.0 & 73.3 & 60.0 &74.7 & 90.0 & 95.0 & 83.3 & 87.5 & 89.3 &89.0 & 3.1 \\
\rowcolor{colorOther}
Qwen-Chat-7B & 66.7 & 73.3 & 46.7 & 53.3 & 33.3 &54.7 & 66.7 & 86.7 & 76.7 & 80.0 & 64.7 &74.9 & 1.6 \\
\rowcolor{colorOther}
LLaMA2-Chat-7B & \textbf{96.7} & \textbf{93.3} & 90.0 & 86.7 & 70.0 &87.3 & \textbf{96.7} & \textbf{96.7} & 95.6 & 96.7 & 93.3 &95.8 & 4.1 \\
\rowcolor{colorOther}
Vicuna-7B-V1.5 & 80.0 & 80.0 & 53.3 & 63.3 & 53.3 &66.0 & 80.0 & 88.3 & 80.0 & 87.5 & 82.0 &83.6 & 2.3 \\
\rowcolor{colorOther}
Baichuan2-Chat-7B & 76.7 & 83.3 & 56.7 & 53.3 & 50.0  &64.0 & 76.7 & 90.0 & 80.0 & 85.8 & 87.3 &84.0 & 2.2 \\
\rowcolor{colorOther}
ChatGLM3-6B  & 80.0&	60.0&	50.0&	36.7&	33.3&	52.0 &80.0&	76.7&	73.3&	74.2&	74.7	&75.8 &1.9
   \\
\bottomrule
\end{tabular}
\caption{Results of Style Constraints across 5 difficulty levels. \colorbox{colorGPT}{Proprietary LLMs}, \colorbox{color70B}{open-sourced LLMs (large)}, \colorbox{color13B}{open-sourced LLMs (medium)}, and \colorbox{colorOther}{open-sourced LLMs (small)} are distinguished by different colors.}
\label{tab:style_level}
\end{table*}

\begin{table*}[!h]
\small
\centering
\setlength{\tabcolsep}{6pt}
\begin{tabular}{l|cccccc|cccccc|c}
\toprule
& \multicolumn{6}{c|}{\textbf{HSR (\%)}} & \multicolumn{6}{c|}{\textbf{SSR (\%)}} \\
\cmidrule(lr){2-7} \cmidrule(lr){8-13}
\multirow{-2}{*}{\centering\textbf{Model}}
& \textbf{L1} & \textbf{L2} & \textbf{L3} & \textbf{L4} & \textbf{L5} & \textbf{Avg.} & \textbf{L1} & \textbf{L2} & \textbf{L3} & \textbf{L4} & \textbf{L5} & \textbf{Avg.}
& \multirow{-2}{*}{\centering\textbf{CSL}} \\
\midrule
\rowcolor{colorGPT}
GPT-4-Preview-1106 & \textbf{90.0} & \textbf{93.3} & \textbf{86.7} & \textbf{93.3} & \textbf{80.0} &\textbf{86.0} & \textbf{90.0} & \textbf{95.0} & \textbf{94.4} & \textbf{98.3} & \textbf{93.3} & \textbf{90.1} & \textbf{4.1} \\
\rowcolor{colorGPT}
GPT-3.5-Turbo-1106 & \textbf{90.0} & 76.7 & 80.0 & 70.0 & 50.0 &73.3 & \textbf{90.0} & 85.0 & 88.9 & 85.0 & 82.0 &86.2 & 3.2 \\
\rowcolor{color70B}
Qwen-Chat-72B & 86.7 &80.0	&83.3	&60.0	&56.7 &73.3 &86.7	&86.7	&87.8	&83.3	&81.3 &85.2 & 3.3\\
\rowcolor{color70B}
LLaMA2-Chat-70B & 83.3	&76.7	&66.7	&53.3	&36.7 &63.3 &83.3	&85.0	&86.7	&78.3	&70.0 &80.7 & 2.4\\
\rowcolor{color13B}
Qwen-Chat-14B & 86.7 &86.7 &80.0	&56.7 &33.3 &68.7 &86.7	&90.0	&86.7	&80.8	&72.0 &83.2 & 3.1\\
\rowcolor{color13B}
WizardLM-13B-V1.2 & 83.3 & \textbf{93.3} & 73.3 & 56.7 & 40.0 &69.3 & 83.3 & \textbf{95.0} & 86.7 & 77.5 & 73.3 &83.2 & 2.9 \\
\rowcolor{color13B}
LLaMA2-Chat-13B & 86.7 & 80.0 & 70.0 & 56.7 & 40.0 &66.7 & 86.7 & 86.7 & 86.7 & 80.0 & 72.7 &82.5 & 3.1 \\
\rowcolor{color13B}
Vicuna-13B-V1.5 & 86.7 & 76.7 & 76.7 & 53.3 & 30.0 &64.7 & 86.7 & 85.0 & 86.7 & 78.3 & 70.7 &81.5 & 2.6 \\
\rowcolor{colorOther}
Qwen-Chat-7B & 76.7 &76.7 &66.7	&50.0 &30.0 &60.0 &76.7	&83.3	&77.8	&74.2	&70.7 &76.5 & 2.4\\
\rowcolor{colorOther}
LLaMA2-Chat-7B & 80.0 & 80.0 & 66.7 & 53.3 & 33.3 &62.7 & 80.0 & 88.3 & 86.7 & 78.3 & 68.0 &80.3 & 2.4 \\
\rowcolor{colorOther}
Vicuna-7B-V1.5 & 80.0 & 76.7 & 73.3 & 43.3 & 20.0 &58.7 & 80.0 & 86.7 & 85.6 & 70.8 & 67.3 &78.1 & 2.4 \\
\rowcolor{colorOther}
Baichuan2-Chat-7B & 80.0 & 56.7 & 60.0 & 40.0 & 36.7 &54.7 & 80.0 & 73.3 & 81.1 & 72.5 & 72.7 &75.9 & 1.9 \\
\rowcolor{colorOther}
ChatGLM3-6B  &  80.0&	60.0&	46.7&	33.3&	23.3&	48.7& 80.0&	71.7&	72.2&	69.2&	68.0 &72.2 &	2.1
  \\
\bottomrule
\end{tabular}
\caption{Results of Format Constraints across 5 difficulty levels. \colorbox{colorGPT}{Proprietary LLMs}, \colorbox{color70B}{open-sourced LLMs (large)}, \colorbox{color13B}{open-sourced LLMs (medium)}, and \colorbox{colorOther}{open-sourced LLMs (small)} are distinguished by different colors.}
\label{tab:format_level}
\end{table*}

\begin{table*}[!h]
\small
\centering
\setlength{\tabcolsep}{6pt}
\begin{tabular}{l|cccccc|cccccc|c}
\toprule
& \multicolumn{6}{c|}{\textbf{HSR (\%)}} & \multicolumn{6}{c|}{\textbf{SSR (\%)}} \\
\cmidrule(lr){2-7} \cmidrule(lr){8-13}
\multirow{-2}{*}{\centering\textbf{Model}}
& \textbf{L1} & \textbf{L2} & \textbf{L3} & \textbf{L4} & \textbf{L5} & \textbf{Avg.} & \textbf{L1} & \textbf{L2} & \textbf{L3} & \textbf{L4} & \textbf{L5} & \textbf{Avg.}
& \multirow{-2}{*}{\centering\textbf{CSL}} \\
\midrule
\rowcolor{colorGPT}
GPT-4-Preview-1106 & \textbf{87.5} & \textbf{57.5} & \textbf{57.5} & \textbf{45.0} & \textbf{42.5} &\textbf{58.0} & \textbf{87.5} & \textbf{57.5} & \textbf{57.5} & \textbf{45.0} & \textbf{42.5} &\textbf{58.0} & \textbf{2.4} \\
\rowcolor{colorGPT}
GPT-3.5-Turbo-1106 & 80.0 & 50.0 & 50.0 & 42.5 & \textbf{42.5} &53.0 & 80.0 & 50.0 & 50.0 & 42.5 & \textbf{42.5} &53.0  & 2.2 \\
\rowcolor{color70B}
Qwen-Chat-72B & 30.0	& 10.0	& 5.0	& 2.5	& 2.5 &10.0 & 30.0	& 10.0	& 5.0	& 2.5	& 2.5 &10.0 & 0.5 \\
\rowcolor{color70B}
LLaMA2-Chat-70B & 0.0	& 2.5	& 0.0	& 0.0	& 0.0 &0.5 & 0.0	& 2.5	& 0.0	& 0.0	& 0.0 &0.5 & 0.0 \\
\rowcolor{color13B}
Qwen-Chat-14B & 22.5	& 10.0	& 5.0	& 2.5	& 7.5 &9.5 & 22.5	& 10.0	& 5.0	& 2.5	& 7.5 &9.5 & 0.4 \\
\rowcolor{color13B}
WizardLM-13B-V1.2 & 40.0 & 30.0 & 27.5 & 12.5 & 15.0 &25.0 & 40.0 & 30.0 & 27.5 & 12.5 & 15.0 &25.0 & 0.9 \\
\rowcolor{color13B}
LLaMA2-Chat-13B & 0.0 & 0.0 & 0.0 & 0.0 & 0.0 & 0.0 & 0.0 & 0.0 & 0.0 & 0.0 & 0.0 & 0.0 & 0.0 \\
\rowcolor{color13B}
Vicuna-13B-V1.5 & 57.5 & 37.5 & 25.0 & 17.5 & 17.5 &31.0 & 57.5 & 37.5 & 25.0 & 17.5 & 17.5 &31.0 & 1.2 \\
\rowcolor{colorOther}
Qwen-Chat-7B & 12.5	& 10.0	& 5.0	& 5.0	& 2.5 &7.0 & 12.5	& 10.0	& 5.0	& 5.0	& 2.5 &7.0 & 0.2 \\
\rowcolor{colorOther}
LLaMA2-Chat-7B & 0.0 & 0.0 & 0.0 & 0.0 & 0.0 & 0.0 & 0.0 & 0.0 & 0.0 & 0.0 & 0.0 & 0.0 & 0.0 \\
\rowcolor{colorOther}
Vicuna-7B-V1.5 & 52.5 & 32.5 & 25.0 & 12.5 & 15.0 &27.5 & 52.5 & 32.5 & 25.0 & 12.5 & 15.0 &27.5 & 1.2 \\
\rowcolor{colorOther}
Baichuan2-Chat-7B & 50.0 & 30.0 & 35.0 & 12.5 & 12.5 &28.0 & 50.0 & 30.0 & 35.0 & 12.5 & 12.5 &28.0 & 1.1 \\
\rowcolor{colorOther}
ChatGLM3-6B  &  32.5&	22.5&	15.0&	10.0&	7.5&	17.5 &32.5&	22.5&	15.0&	10.0&	7.5&	17.5&0.6
  \\
\bottomrule
\end{tabular}
\caption{Results of Example Constraints across 5 difficulty levels. \colorbox{colorGPT}{Proprietary LLMs}, \colorbox{color70B}{open-sourced LLMs (large)}, \colorbox{color13B}{open-sourced LLMs (medium)}, and \colorbox{colorOther}{open-sourced LLMs (small)} are distinguished by different colors.}
\label{tab:example_level}
\end{table*}

\begin{table*}[!h]
\small
\centering
\setlength{\tabcolsep}{6pt}
\begin{tabular}{l|cccccc|cccccc|c}
\toprule
& \multicolumn{6}{c|}{\textbf{HSR (\%)}} & \multicolumn{6}{c|}{\textbf{SSR (\%)}} \\
\cmidrule(lr){2-7} \cmidrule(lr){8-13}
\multirow{-2}{*}{\centering\textbf{Model}}
& \textbf{L1} & \textbf{L2} & \textbf{L3} & \textbf{L4} & \textbf{L5} & \textbf{Avg.} & \textbf{L1} & \textbf{L2} & \textbf{L3} & \textbf{L4} & \textbf{L5} & \textbf{Avg.}
& \multirow{-2}{*}{\centering\textbf{CSL}} \\
\midrule
\rowcolor{colorGPT}
GPT-4-Preview-1106 & 60.0 & 46.7 & 40.0 & \textbf{66.7} & \textbf{40.0} &\textbf{50.7} & 60.0 & 50.0 & 48.9 & \textbf{66.7} & \textbf{56.0} &\textbf{56.3} & \textbf{1.9} \\
\rowcolor{colorGPT}
GPT-3.5-Turbo-1106 & \textbf{70.6} & \textbf{47.1} & \textbf{47.1} & 41.2 & 23.5 &45.9 & \textbf{70.6} & \textbf{52.9} & \textbf{58.8} & 52.9 & 41.2 &55.3 & 1.7 \\
\rowcolor{color70B}
Qwen-Chat-72B &70.6	&52.9	&41.2	&35.3	&17.7 &43.5 &70.6	&55.9	&49.0	&42.7	&38.8 &51.4 &1.8 \\
\rowcolor{color70B}
LLaMA2-Chat-70B &58.8	&35.3	&17.7	&23.5	&17.7 &30.6 &58.8	&41.2	&35.3	&38.2	&37.7 &42.2 &1.2 \\
\rowcolor{color13B}
Qwen-Chat-14B &58.8	&35.3	&35.3	&23.5	&11.8 &32.9 &58.8	&44.1	&41.2	&38.2	&37.7 &44.0 &1.4 \\
\rowcolor{color13B}
WizardLM-13B-V1.2 & 60.0 & 46.7 & 20.0 & 13.3 & 26.7 &33.3 & 60.0 & 46.7 & 37.8 & 36.7 & 41.3 &44.5 & 1.4 \\
\rowcolor{color13B}
LLaMA2-Chat-13B & 47.1 & 41.2 & 35.3 & 29.4 & 29.4 &36.5 & 47.1 & 47.1 & 45.1 & 44.1 & 43.5 &45.4 & 1.5 \\
\rowcolor{color13B}
Vicuna-13B-V1.5 & 64.7 & 41.2 & 29.4 & 23.5 & 23.5 &36.5 & 64.7 & 47.1 & 45.1 & 42.6 & 44.7 &48.9 & 1.5 \\
\rowcolor{colorOther}
Qwen-Chat-7B &64.7	&35.3	&23.5	&17.6	&0.0 &28.2 &64.7	&41.2	&37.3	&33.8	&30.6 &41.5 &1.2 \\
\rowcolor{colorOther}
LLaMA2-Chat-7B & 58.8 & 41.2 & 29.4 & 29.4 & 17.6 &35.3 & 58.8 & 47.1 & 41.2 & 39.7 & 35.3 &44.4 & 1.5 \\
\rowcolor{colorOther}
Vicuna-7B-V1.5 & 47.1 & 29.4 & 17.6 & 17.6 & 11.8 &24.7 & 47.1 & 38.2 & 37.2 & 33.8 & 36.5 &38.6 & 1.1 \\
\rowcolor{colorOther}
Baichuan2-Chat-7B & 47.1 & 17.6 & 11.8 & 17.6 & 11.8 &21.2 & 47.1 & 32.4 & 33.3 & 36.8 & 36.5 &37.2 & 0.7 \\
\rowcolor{colorOther}
ChatGLM3-6B  &  41.2&	29.4&	23.5&	23.5&	17.7 &27.1&	41.2&	38.2&	41.2&	39.7&	35.3 &39.1 &	1.1
  \\
\bottomrule
\end{tabular}
\caption{Results of Mixed Constraints across 5 difficulty levels. \colorbox{colorGPT}{Proprietary LLMs}, \colorbox{color70B}{open-sourced LLMs (large)}, \colorbox{color13B}{open-sourced LLMs (medium)}, and \colorbox{colorOther}{open-sourced LLMs (small)} are distinguished by different colors.}
\label{tab:mixed_level}
\end{table*}

\section{Case Study}
\label{appendix: case_study}

Table \ref{tab:case1} and Table \ref{tab:case2} show the respective responses and evaluation results of GPT-4 and WizardLM-13B-V1.5 when tasked with a level-5 instruction under the category of Content Constraints.
It can be observed that GPT-4 meets all five specified constraints, whereas WizardLM-13B-V1.5 fails to fulfill the third constraint, which mandates that the output animals must be able to swim.
Besides, these two cases also validate the effectiveness of our model-based evaluation.

\begin{table*}[]
\small
\centering
\renewcommand{\arraystretch}{1}
\begin{tabular}{l|l}
\toprule
\multicolumn{2}{c}{\textbf{The response and evaluation result of GPT-4-Preview-1106 for a level-5 instruction}} \\ \toprule
\textbf{Instruction}                        & \parbox[c]{13cm}{
Talk about domestic animal: \\
== the norwegian elkhound is one of the northern spitz type breed of dog and is the national dog of norway. == \\
== the irish terrier (irish: brocaire rua) is a dog breed from ireland, one of many breeds of terrier. == \\
== a hearing dog is a type of assistance dog specifically selected and trained to assist people who are deaf or hard of hearing by alerting their handler to important sounds, such as doorbells, smoke alarms, ringing telephones, or alarm clocks. == \\
Could you list three more domestic animals that satisfy ALL of following conditions: \\
1. Do not talk about dog. \\
2. Their weights are less than 100kg. \\
3. Can swim. \\
4. Have 4 legs. \\
5. Are hairy.
}     \\
\midrule
\textbf{Response}               &  
\parbox[c]{13cm}{
1. The Maine Coon cat is a large domesticated cat breed that originates from the United States. Known for their size, they typically weigh less than 100kg. They have a thick, water-resistant fur coat, which allows them to swim if necessary. \\ \\
2. The American Mink is a semi-aquatic, fur-bearing mammal that is often domesticated for its fur. It has four legs, can swim, and weighs significantly less than 100kg. \\ \\
3. The Norwegian Forest Cat, also known as the "Wegie", is a breed of domestic cat originating in Northern Europe. This breed is adapted to a very cold climate, with a top coat of glossy, long, water-shedding hairs, and a woolly undercoat for insulation. Despite their large size, they weigh less than 100kg and are known to be good swimmers.
}
\\ \midrule
\textbf{Evaluation} &  
\parbox[c]{13cm}{
1) The 5 added constraints are:\\
   - Do not talk about dog.\\
   - Their weights are less than 100kg.\\
   - Can swim.\\
   - Have 4 legs.\\
   - Are hairy.\\\\
2) For the 5 added constraints, the \#Answer of Initial Instruction + 5 constraints\# satisfies each constraint as follows:\\
   - Do not talk about dog: Yes, the answer talks about Maine Coon cat, American Mink, and Norwegian Forest Cat, not dogs.\\
   - Their weights are less than 100kg: Yes, all the animals mentioned weigh less than 100kg.\\
   - Can swim: Yes, all the animals mentioned can swim.\\
   - Have 4 legs: Yes, all the animals mentioned have 4 legs.\\
   - Are hairy: Yes, all the animals mentioned are hairy.\\\\
3) ['YES', 'YES', 'YES', 'YES', 'YES']
}
\\ \bottomrule
\end{tabular}
\caption{The response and evaluation result of GPT-4-Preview-1106 for a level-5 instruction belonging to Content Constraints.}
\label{tab:case1}
\renewcommand{\arraystretch}{1}
\end{table*}

\begin{table*}[]
\small
\centering
\renewcommand{\arraystretch}{1}
\begin{tabular}{l|l}
\toprule
\multicolumn{2}{c}{\textbf{The response and evaluation result of WizardLM-13B-V1.2 for a level-5 instruction}} \\ \toprule
\textbf{Instruction}                        & \parbox[c]{13cm}{
Talk about domestic animal: \\
== the norwegian elkhound is one of the northern spitz type breed of dog and is the national dog of norway. == \\
== the irish terrier (irish: brocaire rua) is a dog breed from ireland, one of many breeds of terrier. == \\
== a hearing dog is a type of assistance dog specifically selected and trained to assist people who are deaf or hard of hearing by alerting their handler to important sounds, such as doorbells, smoke alarms, ringing telephones, or alarm clocks. == \\
Could you list three more domestic animals that satisfy ALL of following conditions: \\
1. Do not talk about dog. \\
2. Their weights are less than 100kg. \\
3. Can swim. \\
4. Have 4 legs. \\
5. Are hairy.
}     \\
\midrule
\textbf{Response}               &  
\parbox[c]{13cm}{
1. Cat: Cats are domestic animals that weigh less than 100kg, have four legs, and are hairy. They are also known for their swimming abilities, making them suitable for households with access to water.\\
2. Rabbit: Rabbits are small, furry animals that weigh less than 100kg and have four legs. They are also excellent swimmers, thanks to their powerful hind legs and water-resistant fur.\\
3. Hamster: Hamsters are small, furry rodents that weigh less than 100kg and have four legs. They are popular pets due to their gentle nature and low maintenance care.
}
\\ \midrule
\textbf{Evaluation} &  
\parbox[c]{13cm}{
1) The 5 added constraints are:\\
   - Do not talk about dog.\\
   - Their weights are less than 100kg.\\
   - Can swim.\\
   - Have 4 legs.\\
   - Are hairy.\\\\
2) For the 5 added constraints:\\
   - Do not talk about dog: The answer does not mention any dogs. It talks about cats, rabbits, and hamsters. So, it satisfies this constraint.\\
   - Their weights are less than 100kg: The animals mentioned in the answer (cats, rabbits, and hamsters) all weigh less than 100kg. So, it satisfies this constraint.\\
   - Can swim: The answer mentions that cats and rabbits can swim. However, hamsters are not natural swimmers. So, it does not satisfy this constraint.\\
   - Have 4 legs: All the animals mentioned in the answer (cats, rabbits, and hamsters) have four legs. So, it satisfies this constraint.\\
   - Are hairy: All the animals mentioned in the answer (cats, rabbits, and hamsters) are described as being hairy or furry. So, it satisfies this constraint.\\\\
3) ['YES', 'YES', 'NO', 'YES', 'YES']
}
\\ \bottomrule
\end{tabular}
\caption{The response and evaluation result of WizardLM-13B-V1.2 for a level-5 instruction belonging to Content Constraints.}
\label{tab:case2}
\renewcommand{\arraystretch}{1}
\end{table*}

\section{\namezh}
\label{appendix: followbench_zh}
To facilitate the multilingual evaluation of LLM's instruction-following ability, we have additionally crafted a Chinese version of \name, denoted as \namezh.
This involved employing a data generation process analogous to that utilized in the development of the English version.
Overall, \namezh consists of 790 meticulously curated instructions from over 50 NLP tasks, including both closed- and open-ended questions. The detailed data statistics are listed in Table \ref{tab:stat_zh}.

\begin{figure}[!t]
\centering
\includegraphics[width=\linewidth]{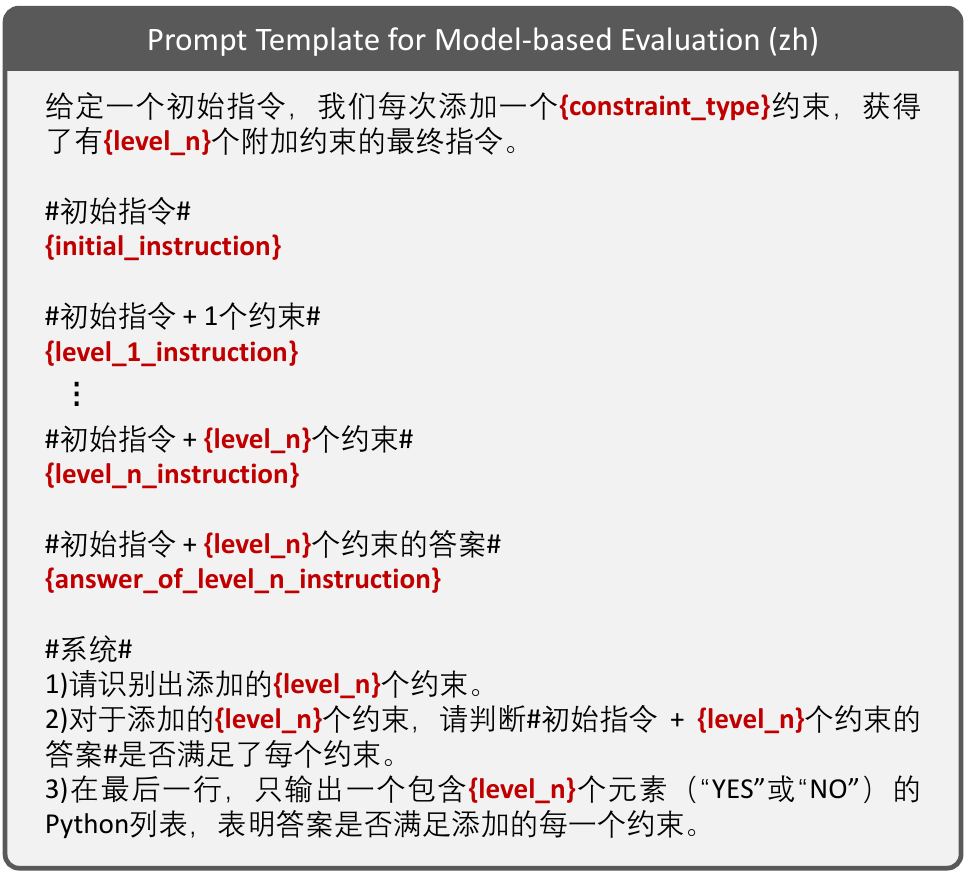}
\caption{
Prompt template for model-based evaluation of \namezh.
}
\label{fig:eval_prompt_zh}
\end{figure}

\begin{figure}[!t]
\centering
\includegraphics[width=\linewidth]{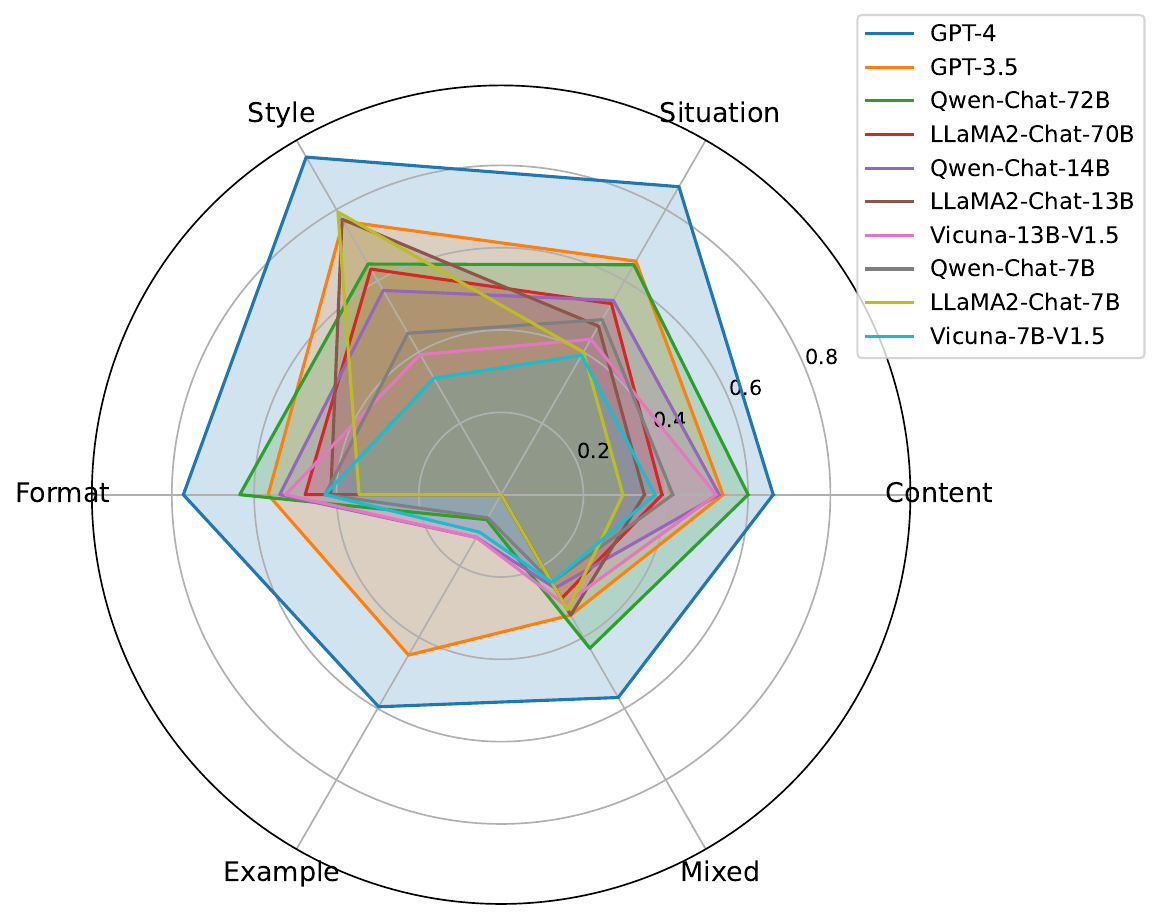}
\caption{
HSR (\%) results in diverse constraint categories of \namezh. For each category, we compute the average score of all difficulty levels.
}
\label{fig:category_zh}
\end{figure}

Following \S\ref{sec: evaluation} and \S\ref{sec: experimental_setup}, we evaluate 13 popular LLMs on \namezh. The prompt template for model-based evaluation of \namezh is shown in Figure \ref{fig:eval_prompt_zh}.
It is noticeable that although LLaMA2-Chat-70B/13B/7B, WizardLM-13B-V1.2, and Vicuna-13B/7B-V1.5 are not specifically trained on Chinese corpora, they can still understand and respond in Chinese.
Table \ref{tab:level_zh} provides a comprehensive comparison of various models across five difficulty levels, denoted as L1 to L5.
Similar to \name, the performance of nearly all models on \namezh typically diminishes as we progress from L1 to L5.
Nevertheless, GPT-3.5 exhibits a notably diminished proficiency in following instructions on \namezh in comparison to GPT-4, showcasing a more pronounced performance gap than observed on \name.
Moreover, models such as Baichuan2-Chat-7B and ChatGLM3-6B, which are pre-trained on a combination of English and Chinese corpora, demonstrate comparable or even better performance compared to their open-source counterparts.
This highlights the significance of incorporating diverse linguistic datasets in pre-training to enhance the multilingual instruction-following capability of LLMs.
Figure \ref{fig:category_zh} depicts the instruction-following capability of LLMs over different constraint categories, with GPT-4 standing out notably among its counterparts.
In a nutshell, there is still a substantial opportunity for enhancing the instruction-following capabilities of existing LLMs.

\begin{table*}[!t]
\footnotesize
\centering
\begin{tabular}{llccc}
\toprule
\textbf{Constraint}        & \textbf{Task}                                        & \textbf{Avg Len} & \textbf{\#Data} & \textbf{Evaluation} \\ \midrule
                           & Data-to-Text Generation                              & 158               & 25              & \includegraphics[width=0.25cm]{figures/python.png}                    \\
                           & Document-Level Event Argument Extraction             & 1,356              & 15              & \includegraphics[width=0.25cm]{figures/python.png}                    \\
                           & Document-Level Named Entity Recognition            & 652              & 25              &  \includegraphics[width=0.25cm]{figures/python.png}                   \\
                           & Text Generation with Language Constraints            & 167               & 25              & \includegraphics[width=0.25cm]{figures/gpt2.jpg}                    \\
\multirow{-5}{*}{Content}  & Open-ended Question Answering                        & 116               & 25              &  \includegraphics[width=0.25cm]{figures/gpt2.jpg}   \\ \midrule
                           & Suggestion Generation & 139               & 40              & \includegraphics[width=0.25cm]{figures/gpt2.jpg}   \\
                           & Role-playing & 203               & 15              & \includegraphics[width=0.25cm]{figures/gpt2.jpg}                      \\
\multirow{-3}{*}{Situation} & Complex Situation Reasoning                                   & 187               & 55              & \includegraphics[width=0.25cm]{figures/python.png} \\  \midrule
Style                      & Open-ended Question Answering & 120               & 150             & \includegraphics[width=0.25cm]{figures/gpt2.jpg}  \\ \midrule
                     & Text-to-Table Generation & 305               & 30             & \includegraphics[width=0.25cm]{figures/python.png} \\
                           \multirow{-3}{*}{Format} & Open-ended Question Answering & 136               & 120             & \includegraphics[width=0.25cm]{figures/gpt2.jpg}  \\ \midrule
Example                    & 40 diverse NLP tasks                                 & 1,556              & 200             & \includegraphics[width=0.25cm]{figures/python.png} \\ \midrule
                           & Text Editing & 195               & 20              & \includegraphics[width=0.25cm]{figures/python.png} \\
                           & Summarization                                        & 481              & 25              & \includegraphics[width=0.25cm]{figures/python.png}                    \\
                           & Machine Translation                                  & 179               & 10              & \includegraphics[width=0.25cm]{figures/python.png}                    \\
\multirow{-4}{*}{Mixed}    & Story Generation                                     & 56               & 10              & \includegraphics[width=0.25cm]{figures/gpt2.jpg}    \\ \bottomrule
\end{tabular}
\caption{An overview of \namezh. ``Avg Len'' is the average character number of instructions. \includegraphics[width=0.25cm]{figures/python.png} refers to rule-based evaluation, while \includegraphics[width=0.25cm]{figures/gpt2.jpg} refers to model-based evaluation.}
\label{tab:stat_zh}
\end{table*}

\begin{table*}[!t]
\small
\centering
\setlength{\tabcolsep}{6pt}
\begin{tabular}{l|cccccc|cccccc|c}
\toprule
& \multicolumn{6}{c|}{\textbf{HSR (\%)}} & \multicolumn{6}{c|}{\textbf{SSR (\%)}} \\
\cmidrule(lr){2-7} \cmidrule(lr){8-13}
\multirow{-2}{*}{\centering\textbf{Model}}
& \textbf{L1} & \textbf{L2} & \textbf{L3} & \textbf{L4} & \textbf{L5} & \textbf{Avg.} & \textbf{L1} & \textbf{L2} & \textbf{L3} & \textbf{L4} & \textbf{L5} & \textbf{Avg.}
& \multirow{-2}{*}{\centering\textbf{CSL}} \\
\midrule
\rowcolor{colorGPT}
GPT-4-Preview-1106 & \textbf{86.7} & \textbf{83.9} & \textbf{68.7} & \textbf{67.0} & \textbf{61.1} & \textbf{73.5} & \textbf{86.7} & \textbf{84.8} & \textbf{76.0} & \textbf{74.0} & \textbf{71.8} & \textbf{78.6} & \textbf{3.1} \\
\rowcolor{colorGPT}
GPT-3.5-Turbo-1106 & 69.6 & 65.2 & 52.8 & 49.1 & 39.5 & 55.2 & 69.6 & 70.8 & 63.8 & 64.0 & 59.5 & 65.5 & 2.2 \\
\rowcolor{color70B}
Qwen-Chat-72B & 66.8 & 58.9 & 47.3 & 42.8 & 36.5 & 50.5 & 66.8 & 62.3 & 59.1 & 59.1 & 57.9 & 61.0 & 2.1 \\
\rowcolor{color70B}
LLaMA2-Chat-70B & 52.8 & 46.4 & 41.0 & 30.3 & 23.5 & 38.8 & 52.8 & 53.1 & 54.0 & 51.0 & 49.1 & 52.0 & 1.5 \\
\rowcolor{color13B}
Qwen-Chat-14B & 64.0 & 48.3 & 42.2 & 33.6 & 25.9 & 42.8 & 64.0 & 57.2 & 56.5 & 53.7 & 51.2 & 56.5 & 1.6 \\
\rowcolor{color13B}
WizardLM-13B-V1.2 & 55.9 & 46.5 & 37.8 & 29.4 & 19.6 & 37.8 & 55.9 & 50.9 & 51.3 & 50.2 & 47.3 & 51.1 & 1.6 \\
\rowcolor{color13B}
LLaMA2-Chat-13B & 53.3 & 46.0 & 36.1 & 30.6 & 29.5 & 39.1 & 53.3 & 51.9 & 50.4 & 48.3 & 49.0 & 50.6 & 1.6 \\
\rowcolor{color13B}
Vicuna-13B-V1.5 & 56.4 & 43.8 & 36.9 & 32.4 & 22.5 & 38.4 & 56.4 & 53.0 & 52.0 & 52.2 & 46.5 & 52.0 & 1.5 \\
\rowcolor{colorOther}
Qwen-Chat-7B & 54.2 & 45.6 & 33.2 & 23.1 & 18.9 & 35.0 & 54.2 & 52.9 & 51.2 & 50.6 & 46.4 & 51.0 & 1.3 \\
\rowcolor{colorOther}
LLaMA2-Chat-7B & 54.0 & 44.7 & 37.6 & 21.7 & 21.7 & 35.9 & 54.0 & 51.3 & 51.0 & 44.2 & 44.4 & 49.0 & 1.5 \\
\rowcolor{colorOther}
Vicuna-7B-V1.5 & 52.6 & 37.8 & 30.0 & 22.0 & 13.4 & 31.2 & 52.6 & 48.8 & 46.6 & 46.7 & 40.5 & 47.1 & 1.2 \\
\rowcolor{colorOther}
ChatGLM3-6B & 62.0 & 45.9 & 36.6 & 28.1 & 17.8 & 38.1 & 62.0 & 53.4 & 54.3 & 49.1 & 45.6 & 52.9 & 1.5 \\
\bottomrule
\end{tabular}
\caption{Results across five difficulty levels of \namezh. For each level, we compute the average score of all constraint categories. \colorbox{colorGPT}{Proprietary LLMs}, \colorbox{color70B}{open-sourced LLMs (large)}, \colorbox{color13B}{open-sourced LLMs (medium)}, and \colorbox{colorOther}{open-sourced LLMs (small)} are distinguished by different colors.}
\label{tab:level_zh}
\end{table*}

\end{document}